\documentclass[10pt,twocolumn,letterpaper]{article}

\usepackage{cvpr}

\usepackage{microtype}

\graphicspath{{./Figures/}}

\renewcommand{\paragraph}[1]{\vspace{.5em}\noindent\textbf{#1.}}

\usepackage{amsmath}
\usepackage[T1]{fontenc}
\usepackage{lmodern}
\usepackage{multirow}
\usepackage[table]{xcolor}
\usepackage{booktabs}
\usepackage{arydshln}
\usepackage{graphicx}

\definecolor{Gray}{gray}{0.9}

\definecolor{cvprblue}{rgb}{0.21,0.49,0.74}
\usepackage[pagebackref,breaklinks,colorlinks,allcolors=cvprblue]{hyperref}

\def\paperID{na} 
\def\confName{CVPR Findings}
\def\confYear{2026}

\title{Hold-One-Shot-Out (HOSO) for Validation-Free Few-Shot CLIP Adapters}

\author{
Chris Vorster, 
Mayug Maniparambil, 
Noel E. O'Connor,
Noel Murphy, 
Derek Molloy \\
\textit{ML-Labs, Dublin City University, Dublin, Ireland} \\
}

\begin{document}
\maketitle
\begin{abstract}
In many CLIP adaptation methods, a blending ratio hyperparameter controls the trade-off between general pretrained CLIP knowledge and the limited, dataset-specific supervision from the few-shot cases. Most few-shot CLIP adaptation techniques report results by ablation of the blending ratio on the test set or require additional validation sets to select the blending ratio per dataset, and thus are not strictly few-shot. We present a simple, validation-free method for learning the blending ratio in CLIP adaptation. Hold-One-Shot-Out (HOSO) presents a novel approach for CLIP-Adapter-style methods to compete in the newly established validation-free setting. CLIP-Adapter with HOSO (HOSO-Adapter) learns the blending ratio using a one-shot, hold-out set, while the adapter trains on the remaining few-shot support examples. Under the validation-free few-shot protocol, HOSO-Adapter outperforms the CLIP-Adapter baseline by more than 4 percentage points on average across 11 standard few-shot datasets. Interestingly, in the 8- and 16-shot settings, HOSO-Adapter outperforms CLIP-Adapter even with the optimal blending ratio selected on the test set. Ablation studies validate the use of a one-shot hold-out mechanism, decoupled training, and improvements over the naively learnt blending ratio baseline. Code is released \href{https://github.com/chris-vorster/HOSO-Adapter}{here}. 
\end{abstract}

\section{Introduction}

Contrastive Image-Language Pre-training (CLIP)  \cite{LearningTransferableVisual2021_radford} introduced a new paradigm for learning visual representations from large-scale image--text pairs, achieving strong downstream performance through zero-shot knowledge transfer. To improve CLIP's adaptability, recent methods fine-tune additional learnable modules, often implemented as lightweight transfer components such as prompt tuning or adapters. Prompt tuning requires end-to-end backpropagation. Adapters update only a small number of parameters, enabling efficient adaptation and improving generalisation to new domains and tasks. Nevertheless, improving few-shot performance without overfitting remains challenging, as these approaches can be highly sensitive to hyperparameters and may depend on large test sets for calibration \cite{CloserLookFewShot2024_silva-rodriguez}. 

\begin{figure}[t]
  \centering
  \includegraphics[width=\columnwidth]{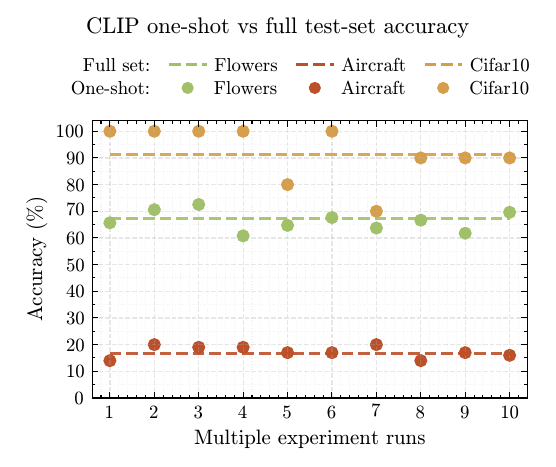}
  \caption{CLIP 1-shot versus full test-set accuracy is strongly correlated across multiple runs. Inspired by this, we use a single hold-out shot to learn the blending ratio for CLIP adapters.}
  \label{fig:one_per_class_runs_combined}
\end{figure}

A central hyperparameter in the few-shot adaptation of vision-language models is the blending ratio. Many CLIP-Adapter-style methods \cite{CLIPAdapterBetterVisionLanguage2025_gao, TipAdapterTrainingfreeAdaption2022_zhang, ProtoAdapterEfficientTrainingFree2024_kato, RCATRetentiveCLIP2024_xie} use a blending ratio, $\alpha$, as the primary parameter for transferring knowledge from the few-shot set while preserving the foundation model's zero-shot knowledge.

\begin{figure}[t]
  \centering
  \includegraphics[width=\columnwidth]{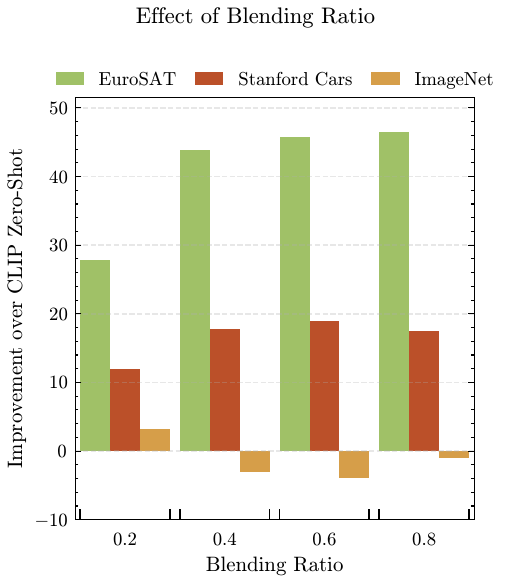}
  \caption{\textbf{The optimal blending ratio reflects a dataset-specific trade-off} between the CLIP prior and task adaptation. Fine-grained datasets like Stanford Cars benefit from a higher $\alpha$ to learn new features, while the general-domain ImageNet performs better with a lower $\alpha$ that preserves the strong prior. This variability makes any fixed ratio suboptimal, motivating our adaptive, validation-free approach.}
  \label{fig:ratio_variability_plot}
\end{figure}

Both the prior and the blending ratio have interesting properties. The prior's performance is indicated by CLIP's zero-shot performance on a domain-specific dataset. For the prior, we observe a strong correlation between CLIP’s zero-shot performance on the full test set and its performance with only a single example per class, as shown in Fig.~\ref{fig:one_per_class_runs_combined}. Furthermore, a single blending ratio value is not optimal across datasets. Fig.~\ref{fig:ratio_variability_plot} depicts the variability in performance due to different blending ratio choices per dataset.  

Since zero-shot performance on a single example per class is strongly correlated with full-test-set performance, we propose using a single instance per class to learn the optimal blending ratio for each dataset.  

In this work, we focus on enabling existing methods with blending ratios to operate under the strict few-shot, validation-free setting recently proposed by Silva-Rodriguez \etal \cite{CloserLookFewShot2024_silva-rodriguez}. In the validation-free setting, methods that use a blending ratio may not use a validation set to select the best ratio per dataset, and hence their performance is severely hampered. \textit{Hold-One-Shot-Out (HOSO)}, a simple approach that learns the blending ratio using a single hold-out shot from the few-shot cases, enables CLIP-Adapter to operate under the strict few-shot protocol even with higher performance in the 8- and 16-shot cases.   

Our contributions are as follows:
\begin{enumerate}
\item \textbf{We introduce Hold-One-Shot-Out (HOSO)}, a novel, validation-free strategy for learning the blending ratio in adapter-based models.
\item Our implementation, \textbf{HOSO-Adapter}, sets a new state-of-the-art for validation-free blending ratio learning in CLIP-Adapter-style methods, outperforming existing baselines by up to 4 points on average. Critically, it even surpasses the performance of a fixed grid-searched oracle in higher-shot settings.
\item \textbf{We provide a controlled evaluation} of learnable blending ratios by reimplementing and isolating key components from prior work, establishing fair and reproducible baselines for this specific task.
\item We provide a \textbf{comprehensive empirical analysis} that validates our key design principles. We demonstrate the necessity of a single-shot cache, decoupled optimisation, and we show that with HOSO, the blending ratio acts as a dynamic regulariser to prevent overfitting.
\end{enumerate}

\section{Related Work}

Few-shot CLIP adaptation can be divided into prompt learning and adapter-based approaches. Prompt learning optimises global text or visual prompts for downstream tasks but requires computationally expensive backpropagation through the text encoder \cite{LearningPromptVisionLanguage2022_zhoua, ConditionalPromptLearning2022_zhoua, MaPLeMultimodalPrompt2023_khattak}. Adapter-based methods operate in feature space and do not require access to the pre-trained weights. Training-based adapters either fit a linear classifier \cite{TransferringVisionLanguageModels2024_wu, LPSurprisinglyStrong2024_huang} or train a shallow MLP \cite{CLIPAdapterBetterVisionLanguage2025_gao, EnhancingCLIPGPT42023_maniparambil} to adapt CLIP features. Tip-Adapter implements a training-free key-value cache populated from few-shot examples \cite{TipAdapterTrainingfreeAdaption2022_zhang}. The validation-free few-shot protocol fixes hyperparameters across datasets rather than tuning them on a validation set, a setting also explored by linear-probe methods like CLAP \cite{CloserLookFewShot2024_silva-rodriguez} and ProKeR~\cite{ProKeRKernelPerspective2025_bendou}. RCAT combines CLIP with a Retentive Network for few-shot video recognition and uses a specialised adapter to align CLIP with temporal and spatial characteristics, reporting a peak blending ratio \(\alpha=0.2\) on UCF101 \cite{UCF101Dataset1012012_soomro} via grid search with RCAT-L/14 \cite{RCATRetentiveCLIP2024_xie}. Proto-Adapter \cite{ProtoAdapterEfficientTrainingFree2024_kato} uses a single-layer class-prototype adapter and linearly blends adapter logits with CLIP zero-shot logits using a hardcoded mixing coefficient \(\alpha\) per dataset.

The works listed are a subset of all CLIP Adapters that use a grid-searched blending ratio per dataset and thus fall outside the strict few-shot setting. CLIP-Adapter and Proto-Adapter report dataset-dependent optimal values of \(\alpha\): fine-grained benchmarks favour larger \(\alpha\) while generic datasets prefer smaller \(\alpha\); TipAdapter prescribes larger \(\alpha\) when the domain gap is substantial. Please refer to Suppl.~Section A for a more detailed description of related works. 

\paragraph{Learnable blending ratio} To the best of our knowledge, only two works have proposed methods to select the blending ratio of CLIP adapters in a validation-free manner, namely SVL-Adapter \cite{SVLAdapterSelfSupervisedAdapter2022_pantazis} and PathCLIP \cite{BridgingPathologyDomain2025_lai}. 

SVL-Adapter combines the complementary strengths of vision–language pretraining and self‑supervised representation learning, and introduces a fully automatic method for selecting the blending hyperparameter \(\alpha\) without requiring held‑out labelled validation data. The method computes \(\alpha\) from CLIP's average prediction confidence on the \(N\) test images of a dataset: \(\alpha = \frac{1}{N}\sum_{i=1}^N \max_k P(y_i = k \mid x_i)\), which directly reflects CLIP's confidence in its zero‑shot predictions. This mechanism operates under the assumption that when CLIP is not confident, the influence of low‑shot learning should increase, automatically adjusting the contribution of low‑shot learning relative to zero‑shot CLIP. 

PathCLIP derives a dynamic blending ratio from two signals on the support set: (1) representational consistency between weakly and strongly augmented views of the few-shot cases, and (2) standard training accuracy. This combination creates a mechanism to learn the blending ratio validation-free. 

\textbf{Our distinction}. SVL-Adapter computes the blending ratio \(\alpha\) from CLIP's average zero‑shot prediction confidence, providing a confidence‑driven heuristic. PathCLIP adapts CLIP with a self‑adaptive \(\alpha\) by combining a Dual‑view Vision Contrastive metric and training accuracy under bound constraints. We use these components from the two methods as baselines. Our method differs in that it sets \(\alpha\) as a learnable logit and optimises that logit independently of the adapter weights via a hold‑one‑shot‑out cache.  

\begin{table*}[ht]
\caption{\textbf{Transfer learning performance (16-shot, RN50).}}
\label{table_per_dataset_results}
\centering
\resizebox{1\textwidth}{!}{
\begin{tabular}{lccccccccccccc}
\toprule
Method & ImageNet & Caltech101 & OxfordPets & StanfordCars & Flowers102 & Food101 & FGVCAAircraft & SUN397 & DTD & EuroSAT & UCF101 & Average \\
\midrule
CLIP-Adapter$_\text{ IJCV'23}$\cite{CLIPAdapterBetterVisionLanguage2025_gao}$^\dagger$ & 63.59 & 92.49 & 87.84 & 74.01 & 93.90 & 78.25 & 32.10 & 69.55 & 65.96 & 84.43 & 76.76 & 74.44 \\
\hline
Zero-Shot$_\text{ ICML'21}$\cite{LearningTransferableVisual2021_radford} & 60.35 & 83.81 & 82.86 & 55.69 & 65.94 & 74.85 & 17.16 & 56.80 & 42.32 & 37.53 & 57.47 & 57.71 \\
CLIP-Adapter$_\text{ IJCV'23}$\cite{CLIPAdapterBetterVisionLanguage2025_gao} & 59.02 & 92.28 & 84.92 & 73.49 & 94.56 & 73.96 & 34.19 & 68.14 & 65.70 & 83.24 & 77.30 & 73.35 \\
TIP-Adapter$_\text{ ECCV'22}$\cite{TipAdapterTrainingfreeAdaption2022_zhang} & 57.81 & 88.44 & 81.09 & 58.83 & 78.41 & 72.96 & 21.96 & 64.00 & 54.79 & 67.90 & 64.52 & 64.61 \\
SVL-Adapter$_\text{ BMVC'22}$\cite{ SVLAdapterSelfSupervisedAdapter2022_pantazis}$^*$ & -- & 90.00 & \textbf{89.50} & 54.00 & 87.00 & 78.00 & 29.70 & 65.50 & 63.00 & \textbf{97.50} & 70.50 & -- \\
SVL-Adapter$_\text{ BMVC'22}$\cite{ SVLAdapterSelfSupervisedAdapter2022_pantazis}$^\S$ & 51.40 & 82.90 & 71.10 & 35.00 & 81.40 & 38.80 & 26.40 & 59.60 & 53.80 & 75.10 & 63.70 & 58.11 \\
PathCLIP$_\text{ ECCV'24}$\cite{BridgingPathologyDomain2025_lai}$^\ddagger$ & \textbf{63.50} & 92.70 & 84.90 & \textbf{73.80} & 94.10 & 77.50 & 34.30 & \textbf{70.30} & 66.40 & 81.90 & \textbf{79.00} & 73.35 \\
HOSO-Adapter (ours) & 62.93 & \textbf{93.03} & 89.47 & \textbf{73.80} & \textbf{95.07} & \textbf{80.93} & \textbf{34.60} & 69.83 & \textbf{66.77} & 83.27 & 78.03 & \textbf{75.25} \\
\bottomrule
\end{tabular}
}
\vspace{2mm}
\begin{minipage}{\textwidth}

\footnotesize

$^\dagger$ Original CLIP-Adapter paper results; blending ratio chosen per dataset. Included for reference, not directly comparable.\\
$^*$ Estimated blending ratio version of SVL-Adapter. Results from their paper, which did not include ImageNet.\\
$^\S$ SVL-Adapter reimplemented without SSL; i.e. the zero-shot CLIP confidence scores are used as the blending ratio.\\
$^\ddagger$ The PathCLIP authors did not evaluate transfer-learning performance; we reimplement their Dual-view Vision Contrastive method to learn the blending ratio.

\end{minipage}
\end{table*}

\section{Preliminaries}

Large-scale VLMs, such as CLIP~\cite{LearningTransferableVisual2021_radford}, are trained on extensive collections of image-text pairs to learn a shared multimodal representation. CLIP consists of two encoders: a vision encoder $f_{\theta}(\cdot)$ and a text encoder $f_{\phi}(\cdot)$, which produce $\ell_2$-normalised embeddings $v$ and $t$ for an image and a text input, respectively. Pre-training aligns paired image-text embeddings and separates mismatched pairs using a contrastive objective, thereby encouraging semantic consistency across modalities in the joint embedding space. 

For a downstream classification task, CLIP can perform recognition directly without task-specific training. Given $C$ categories and an ensemble of $N$ natural-language prompts for each class, $\{\{T_{n,c}\}_{n=1}^{N}\}_{c=1}^{C}$, a class prototype is computed as the average text embedding:
\begin{equation}
t_c = \frac{1}{N}\sum_{n=1}^{N} f_{\phi}(T_{n,c}).
\end{equation}
For an input image $x$, its visual embedding is $v = f_{\theta}(x)$, and the prediction probabilities over the $C$ classes are obtained using a softmax over cosine similarities:
\begin{equation}
\hat{y}_c = \frac{\exp(v \cdot t_c / \tau)}{\sum_{i=1}^{C} \exp(v \cdot t_i / \tau)},
\label{eq:zeroshot}
\end{equation}
where $\tau$ is the temperature parameter learnt during pre-training, and $v \cdot t_c$ denotes the dot product, equivalent to cosine similarity since all embeddings are $\ell_2$-normalised.

In the few-shot regime, a method has access to a limited labelled support set $\mathcal{S}=\{(x^{(m)},y^{(m)})\}_{m=1}^{M}$ with $K$ examples per class, \ie,\ $M=K\times C$, where $K\in\{1,2,4,8,16\}$. Each label $y^{(m)}\in\{0,1\}^{C}$ corresponds to the class of the image $x^{(m)}$. The objective is to adapt the pre-trained model to the target task using $\mathcal{S}$ while preserving its generalisation ability. Due to the limited supervision, direct fine-tuning of the full model is often infeasible, motivating the use of lightweight adaptation mechanisms. 

To efficiently adapt VLMs to new tasks, adapter methods introduce small trainable modules into the frozen backbone. Let the original visual feature be $v = f_{\theta}(x)$. An adapter module parameterised by $\psi$ transforms this feature as
\begin{equation}
v' = v + g_{\psi}(v),
\end{equation}
where $g_{\psi}(\cdot)$ is a lightweight function, typically implemented as a bottleneck Multi-Layer Perceptron or a residual linear layer. Only the adapter parameters $\psi$ are optimised during fine-tuning, while $\theta$ and $\phi$ remain fixed. This design substantially reduces the number of trainable parameters and prevents catastrophic forgetting \cite{OvercomingCatastrophicForgetting2017_kirkpatrick}, making adapters particularly effective in few-shot or resource-limited scenarios. Moreover, by decoupling task-specific learning from the frozen backbone, adapter-based methods allow efficient reuse of pre-trained representations and facilitate privacy-preserving, black-box adaptation.

\section{Method}

\label{sec:method} 

Building on adapter-based fine-tuning, we introduce a novel, validation-free method that incorporates a learnable blending ratio and a decoupled optimisation strategy. Our approach dynamically adjusts the contributions of the pre-trained CLIP features and the learned task-specific features by optimising this balance on a 1-shot hold-out set. 

Given a visual feature $v = f_{\theta}(x)$ from the frozen CLIP encoder, and an adapted feature $v_{\text{adapt}} = g_{\psi}(v)$ produced by a lightweight bottleneck adapter module, the final image embedding $\hat{v}$ is the linear combination:
\begin{equation}
\hat{v} = (1 - \alpha) \cdot v + \alpha \cdot v_{\text{adapt}},
\label{eq:interpolation}
\end{equation}
where $\alpha \in [0, 1]$ is a learnable blending ratio that controls the blending of the two feature streams. 

To ensure stable optimisation and constrain the interpolation weight to a meaningful range, HOSO-Adapter parametrises $\alpha$ using a learnable logit, $\alpha_{\text{logit}}$, which is transformed via a scaled sigmoid function:
\begin{equation}
\alpha = \text{sigmoid}(\alpha_{\text{logit}}) \cdot 0.8 + 0.1.
\label{eq:alpha_transform}
\end{equation}
This transformation bounds $\alpha$ within the interval $[0.1, 0.9]$, preventing the model from discarding either the original or adapted feature stream entirely. The resulting embedding $\hat{v}$ is then $\ell_2$-normalised before being used for classification, and the final prediction is computed as described in Eq.~\ref{eq:zeroshot}. During training, the CLIP backbone parameters ($\theta$, $\phi$) are frozen, while the adapter parameters $\psi$ and the scalar logit $\alpha_{\text{logit}}$ are optimised.

A key challenge is to optimise both the adapter parameters $\psi$ and the blending ratio $\alpha$ simultaneously without overfitting. To resolve this, we introduce a decoupled optimisation scheme that leverages the proposed \textit{hold-one-shot-out} approach.

\paragraph{Cache creation and text feature pre-computation} At initialisation, we construct a hold-out cache $\mathcal{C}$ by selecting exactly one image-label pair per class from the full $K$-shot support set $\mathcal{S}$. These $C$ samples are subsequently removed from the main few-shot training data, leaving a reduced support set $\mathcal{S}' = \mathcal{S} \setminus \mathcal{C}$. We pre-compute the text features $\{t_c\}_{c=1}^{C}$ for all classes. We use the same single, dataset-specific prompt templates (e.g., ``a photo of a \{\}, a type of flower.'') as CLIP-Adapter. These class prototypes are computed once and stored, eliminating redundant forward passes through the text encoder. 

We employ two distinct optimisers to train the adapter and the blending ratio on separate data partitions.
\begin{itemize}
\item \textbf{Adapter Training:} The adapter parameters $\psi$ are trained to minimise the cross-entropy loss on the main support set $\mathcal{S}'$. This allows the adapter to learn task-relevant features from the available $(K-1)$-shot examples per class.
\item \textbf{Ratio Training:} The interpolation logit $\alpha_{\text{logit}}$ is optimised separately using a dedicated stochastic gradient descent optimiser to minimise cross-entropy loss exclusively on the hold-one-shot-out cache $\mathcal{C}$.
\end{itemize}

The decoupled strategy treats the cache as a micro-validation set for tuning the blending ratio $\alpha$. By creating a hold-one-shot-out cache to train the blending ratio, our method learns both task-specific features and their appropriate contributions to the final representation, even in the highly data-scarce few-shot setting. 



\begin{table*}[ht]
\caption{\textbf{ViT-B/16 16-shot performance comparison.} CLIP-Adapter results are shown with a fixed residual ratio (\(\alpha=0.2\)) and with the best-performing ratio selected per dataset (best $\alpha$). Results are averaged over three runs.}
\label{table:vitb16_comparison}
\centering
\resizebox{\textwidth}{!}{
\begin{tabular}{lcccccccccccr}
\toprule
Method & Caltech101 & DTD & EuroSAT & FGVCAircraft & Food101 & ImageNet & Flowers102 & OxfordPets & StanfordCars & SUN397 & UCF101 & Average \\
\midrule
CLIP-Adapter (best $\alpha$)$^{\dagger}$ & 95.90 & 71.70 & 85.80 & 45.80 & 89.30 & 71.50 & 97.40 & 92.70 & 82.10 & 75.60 & 84.00 & 81.07 \\
\hline
CLIP-Adapter ($\alpha=0.2$)$^{\ast}$ & 94.90 & 59.70 & 70.50 & 34.10 & \textbf{89.10} & \textbf{71.50} & 93.10 & \textbf{92.60} & 73.90 & 74.20 & 80.40 & 75.82 \\
HOSO-Adapter (ours) & \textbf{95.40} & \textbf{70.67} & \textbf{85.30} & \textbf{43.23} & 88.97 & 70.93 & \textbf{97.23} & 92.27 & \textbf{81.50} & \textbf{74.67} & \textbf{83.43} & \textbf{80.33} \\
\bottomrule
\end{tabular}
}
\vspace{2mm}
\begin{minipage}{\textwidth}
\footnotesize
$^{\dagger}$ Best residual ratio from the set \{0.2, 0.4, 0.6, 0.7, 0.8\} selected per dataset, hence not directly comparable, but included for reference.
$^{\ast}$ CLIP-Adapter under the validation-free setting where the optimal $\alpha$ is selected on ImageNet. 
\end{minipage}
\end{table*}

\section{Experimental Setup}
\subsection{Datasets}
Prior work benchmarks methods across 11 diverse datasets, including ImageNet, Caltech101, OxfordPets, StanfordCars, Flowers102, Food101, FGVCAircraft, SUN397, DTD, EuroSAT and UCF101 \cite{deng2009imagenet,caltech,oxfordpets,stanfordcars,flowers102,food101,aircraft,sun397,dtd,eurosat,ucf101}. These datasets cover a wide range of computer vision classification tasks, including general object, action and fine-grained category recognition in both broad and specialised domains. Few-shot adapters are trained by randomly sampling $K$ shots per class, where $K \in \{2,4,8,16\}$. Evaluation uses the test sets and data splits provided by the authors of CLIP-Adapter \cite{CLIPAdapterBetterVisionLanguage2025_gao}. 
\subsection{Implementation Details}

We implement our proposed method in PyTorch and run all experiments on an NVIDIA RTX 4090 GPU. HOSO-Adapter uses a pre-trained CLIP model \cite{LearningTransferableVisual2021_radford} with a ResNet-50 \cite{IMAGEWORTH16X162021_dosovitskiya} vision backbone (unless otherwise specified). We keep all hyperparameters the same as in CLIP-Adapter for fair comparison. The adapter module is a bottleneck MLP with a reduction factor of 4 and ReLU activations. HOSO-Adapter is trained using stochastic gradient descent (SGD) with momentum 0.9, weight decay 0.0005, and an initial learning rate of 0.002, decayed using a cosine annealing schedule. Training runs for 200 epochs with a batch size of 32; standard data augmentations are applied, including random resized cropping to 224×224 and random horizontal flipping. The learnable blending ratio is initialised to 0.5 and optimised separately on the hold-one-shot-out cache with SGD and a learning rate of 0.1. Reported results are averaged over three runs with distinct random seeds to select different few-shot samples. 
\subsection{Baselines}
HOSO-Adapter compares against the only two existing approaches for learning a blending ratio for CLIP adaptation (SVL-Adapter \& PathCLIP) and against a CLIP-Adapter baseline in which all other components and hyperparameters are held constant and only the blending ratio differs. The authors of PathCLIP have not released their code, and they do not report results on the standard few-shot benchmarks; therefore, PathCLIP is reimplemented for this study (see Suppl.~Section B for implementation details). SVL-Adapter comprises multiple components (a learnable blending ratio based on CLIP confidence scores and a self-supervised encoder); both the reported results for the full SVL-Adapter and a reimplementation that isolates only the learnable blending-ratio component are included (see Suppl. Section C). 

\section{Results}

\label{sec:results}

\subsection{Main Results}
We present a comprehensive evaluation of HOSO-Adapter across 11 standard benchmarks. Our experiments cover two CLIP backbone architectures (ResNet-50 and ViT-B/16) and varying few-shot settings ($K \in \{2, 4, 8, 16\}$).

\paragraph{ResNet-50 results}
Table~\ref{table_per_dataset_results} presents a per-dataset breakdown of accuracy (\%) for few-shot adaptation of CLIP with a ResNet-50 backbone. We compare our results against other methods that have blending ratios. We use the reimplementations of CLIP-Adapter \cite{CLIPAdapterBetterVisionLanguage2025_gao} and TIP-Adapter \cite{TipAdapterTrainingfreeAdaption2022_zhang} under the strict few-shot setting by Silva-Rodriguez \etal \cite{CloserLookFewShot2024_silva-rodriguez} and include SVL-Adapter \cite{SVLAdapterSelfSupervisedAdapter2022_pantazis} that complies with the strict setting. The original CLIP-Adapter results, where the best blending ratio is selected per test dataset, are included for reference. Interestingly, HOSO-Adapter outperforms CLIP-Adapter despite identical architecture and hyperparameters; the only difference is the trained blending parameter. 

\paragraph{ViT-B/16 results}
To assess the scalability of our approach, we evaluate HOSO-Adapter using the more powerful ViT-B/16 backbone~\cite{IMAGEWORTH16X162021_dosovitskiya}. As shown in Table~\ref{table:vitb16_comparison}, HOSO-Adapter achieves an average accuracy of 80.33\% in the 16-shot setting, significantly outperforming the validation-free CLIP-Adapter baseline by over 4.5 percentage points. A per-dataset analysis reveals that these gains are most substantial on fine-grained and specialised datasets. For instance, on the challenging Describable Textures (DTD) and FGVCAircraft datasets, HOSO-Adapter improves accuracy by +11.0 and +9.1 points, respectively. The largest improvement is on EuroSAT, with a significant +14.8 point gain. Crucially, our validation-free method nearly closes the performance gap with the oracle CLIP-Adapter, which relies on a per-dataset grid search over the test set (81.07\%). These findings validate that HOSO-Adapter is a backbone-agnostic strategy that effectively regularises few-shot adaptation across both general and fine-grained domains.  

\begin{figure*}[ht]
    \centering
    \begin{minipage}[t]{0.32\linewidth}
    \centering
    \includegraphics[width=2.2in]{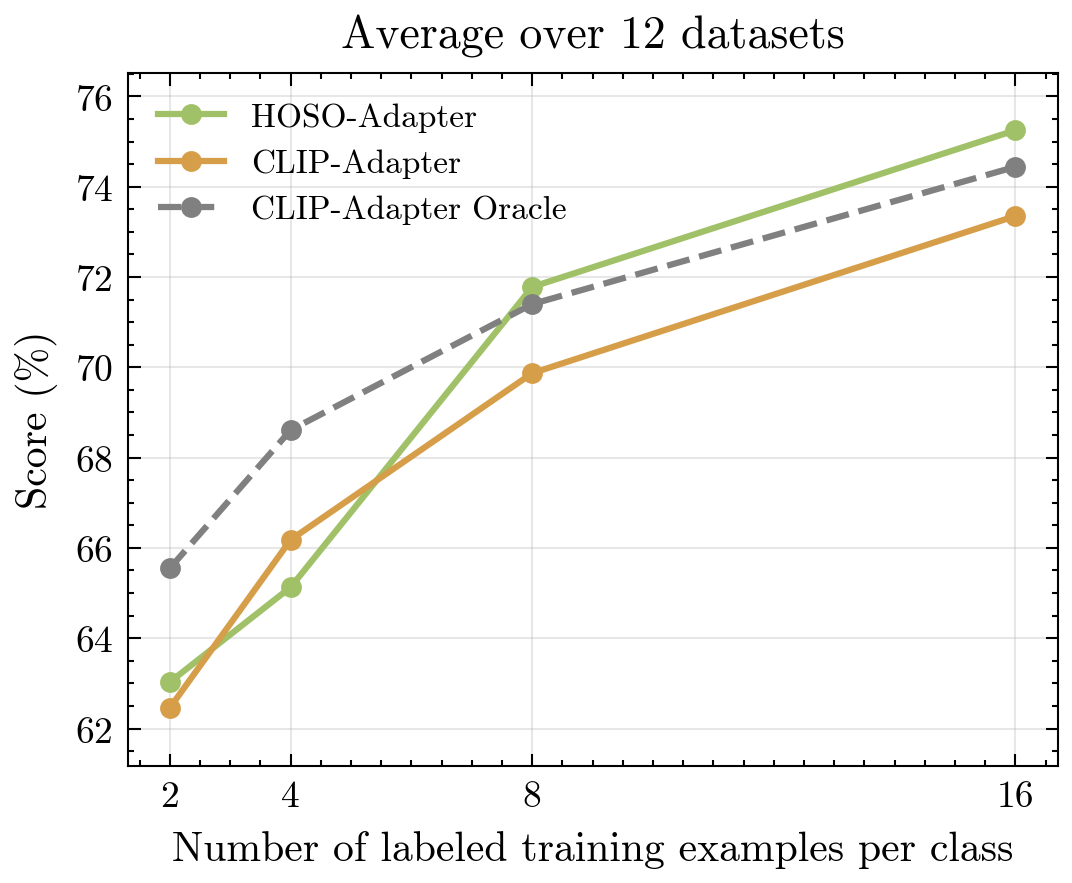}
    \end{minipage}
    \hfill
    \begin{minipage}[t]{0.32\linewidth}
    \centering
    \includegraphics[width=2.2in]{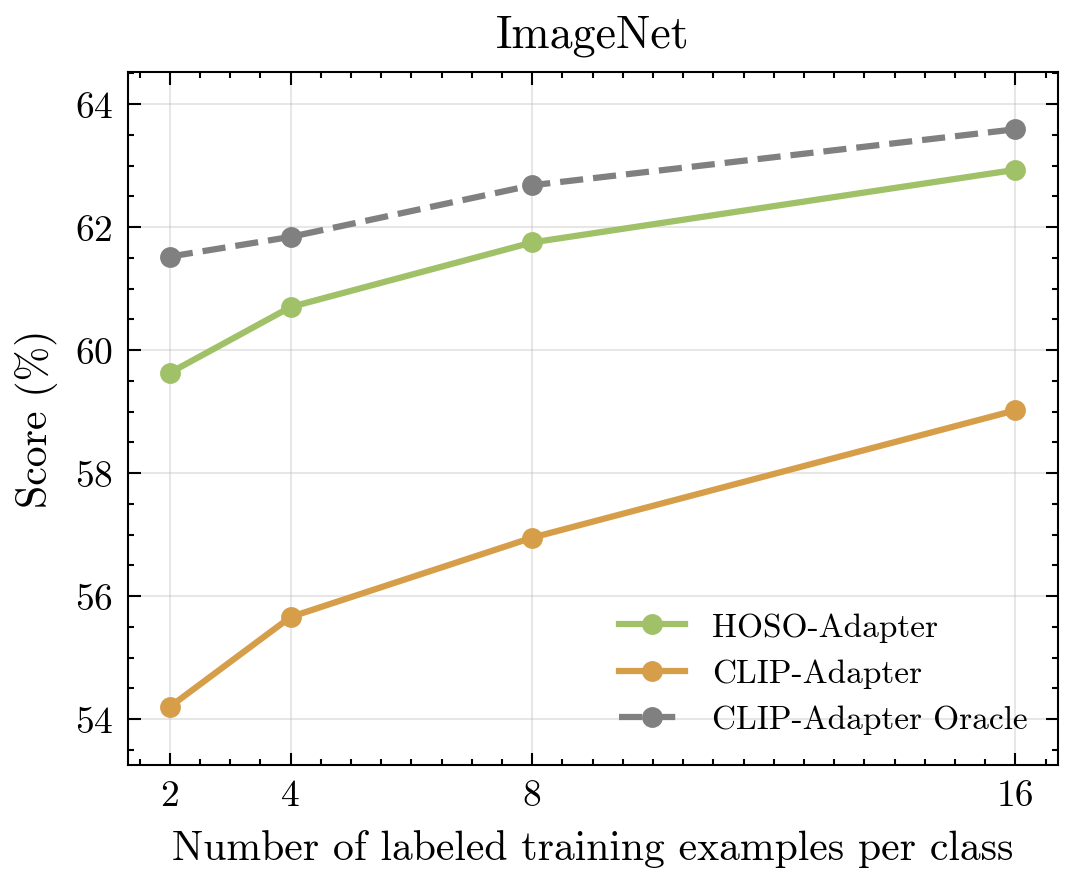}
    \end{minipage}
    \hfill
    \begin{minipage}[t]{0.32\linewidth}
    \centering
    \includegraphics[width=2.2in]{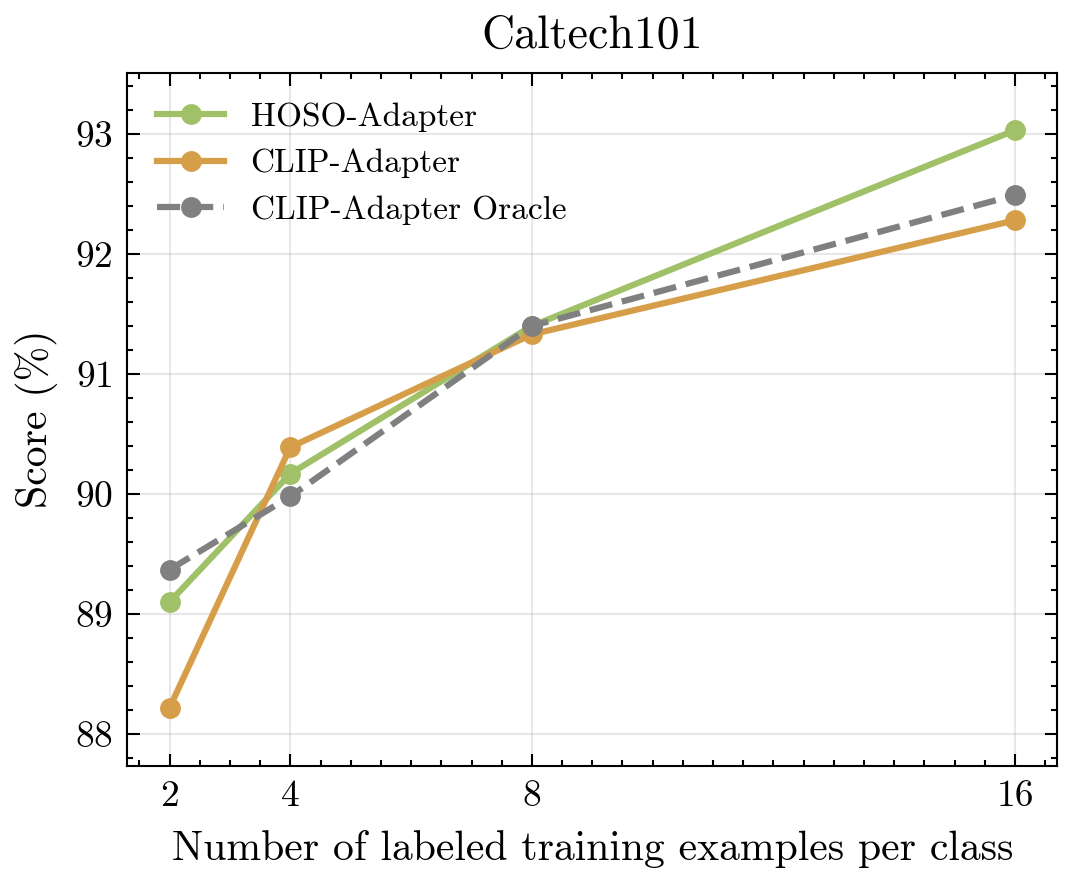}
    \end{minipage}
    
    \vspace{0.05in}
    
    \begin{minipage}[t]{0.32\linewidth}
    \centering
    \includegraphics[width=2.2in]{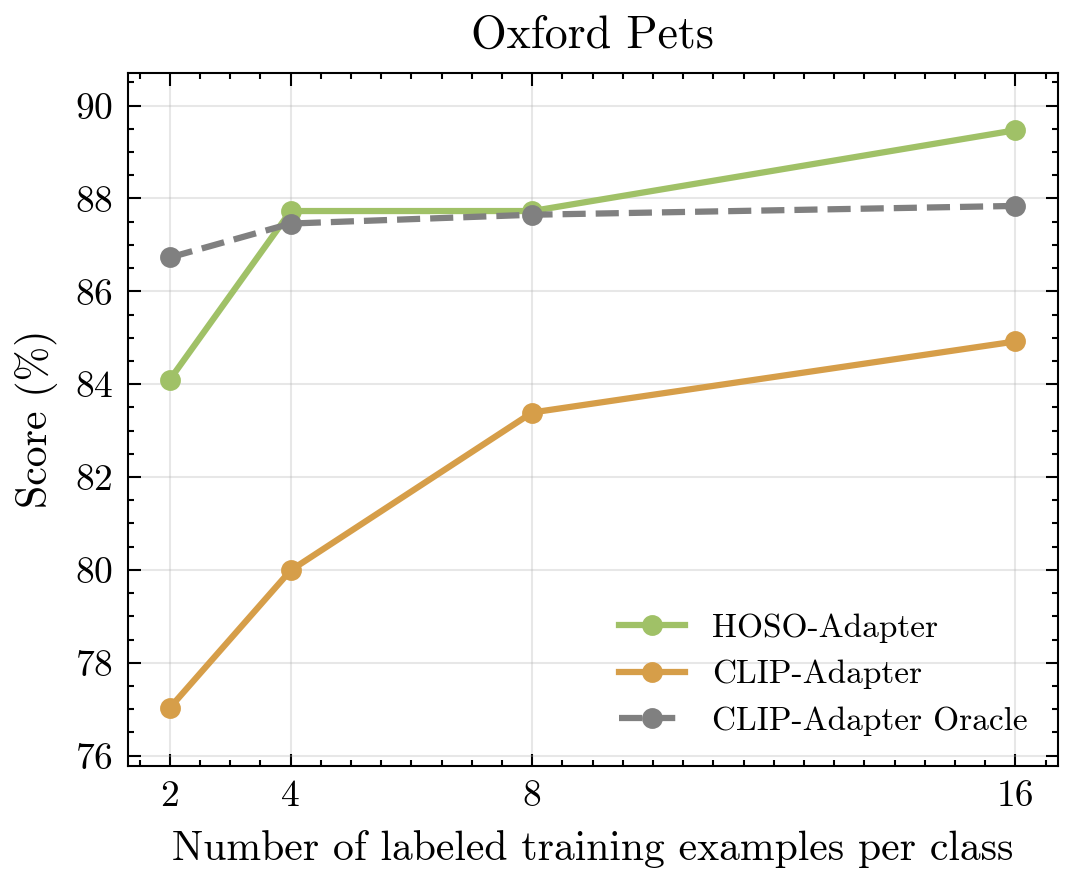}
    \end{minipage}
    \hfill
    \begin{minipage}[t]{0.32\linewidth}
    \centering
    \includegraphics[width=2.2in]{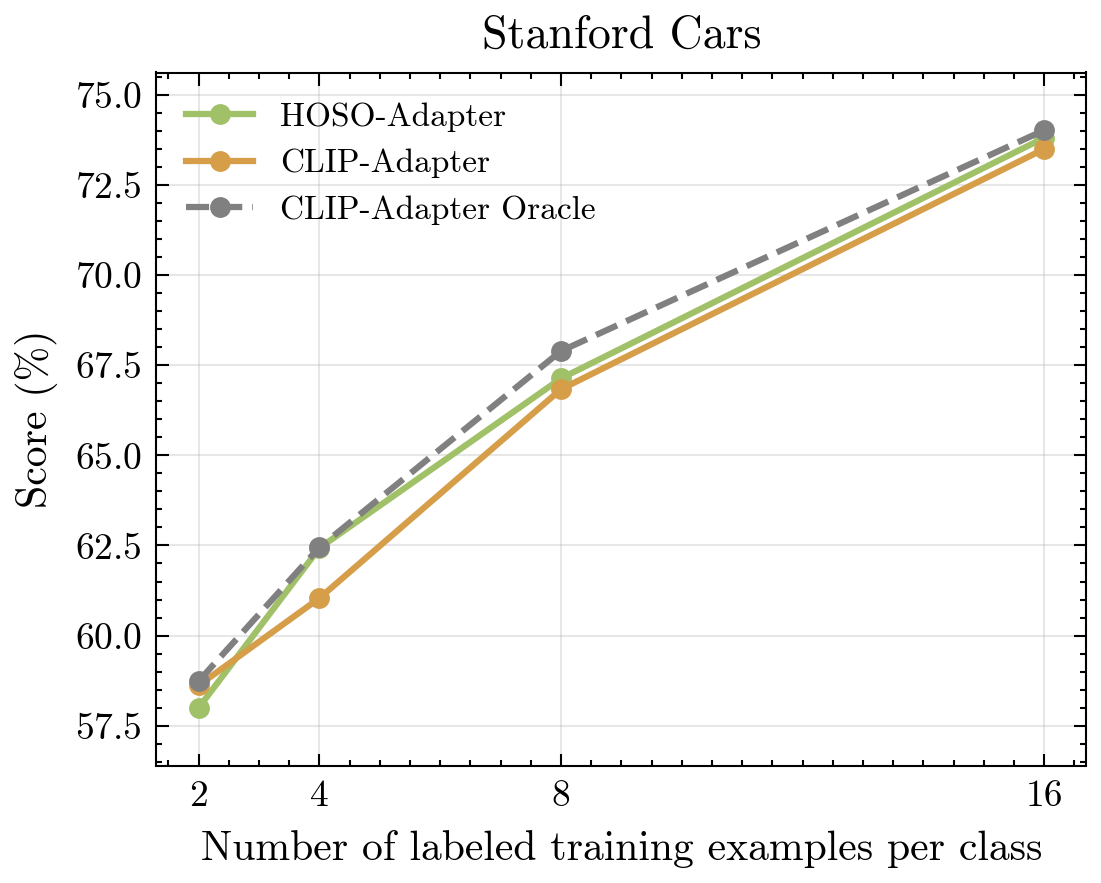}
    \end{minipage}
    \hfill
    \begin{minipage}[t]{0.32\linewidth}
    \centering
    \includegraphics[width=2.2in]{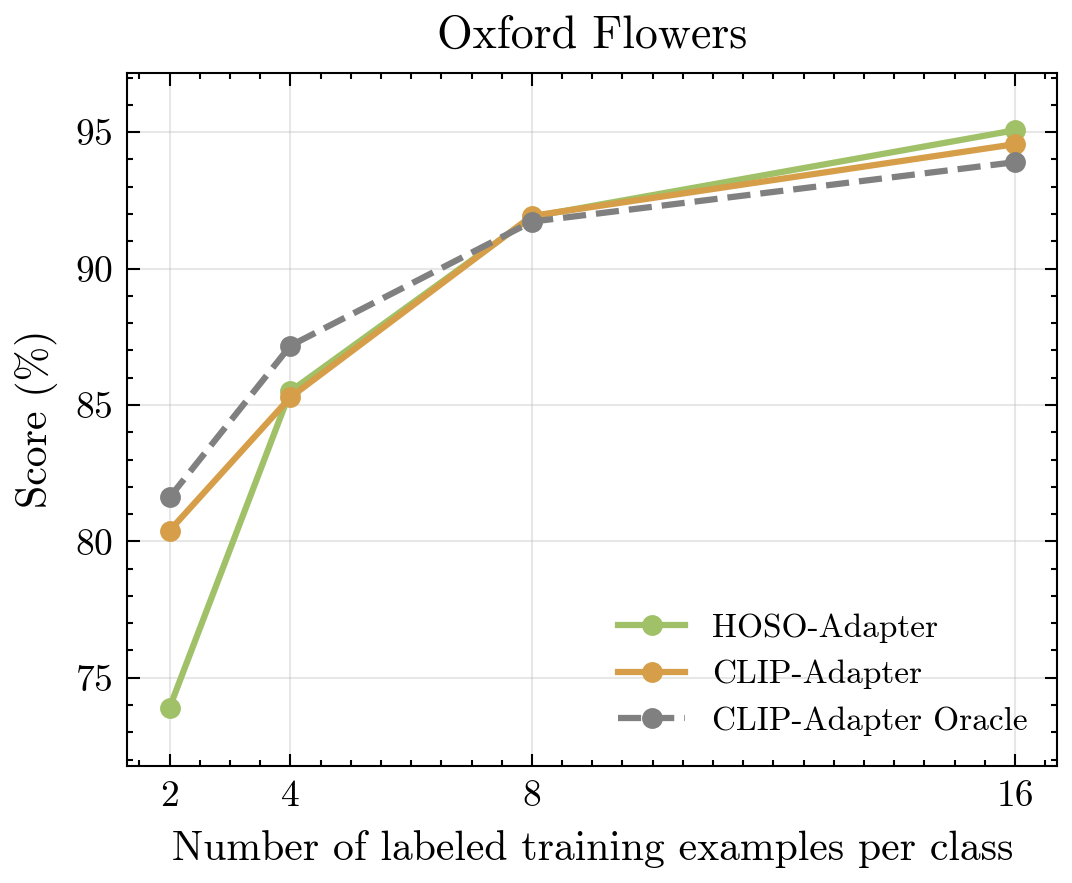}
    \end{minipage}
    
    \vspace{0.05in}
    
    \begin{minipage}[t]{0.32\linewidth}
    \centering
    \includegraphics[width=2.2in]{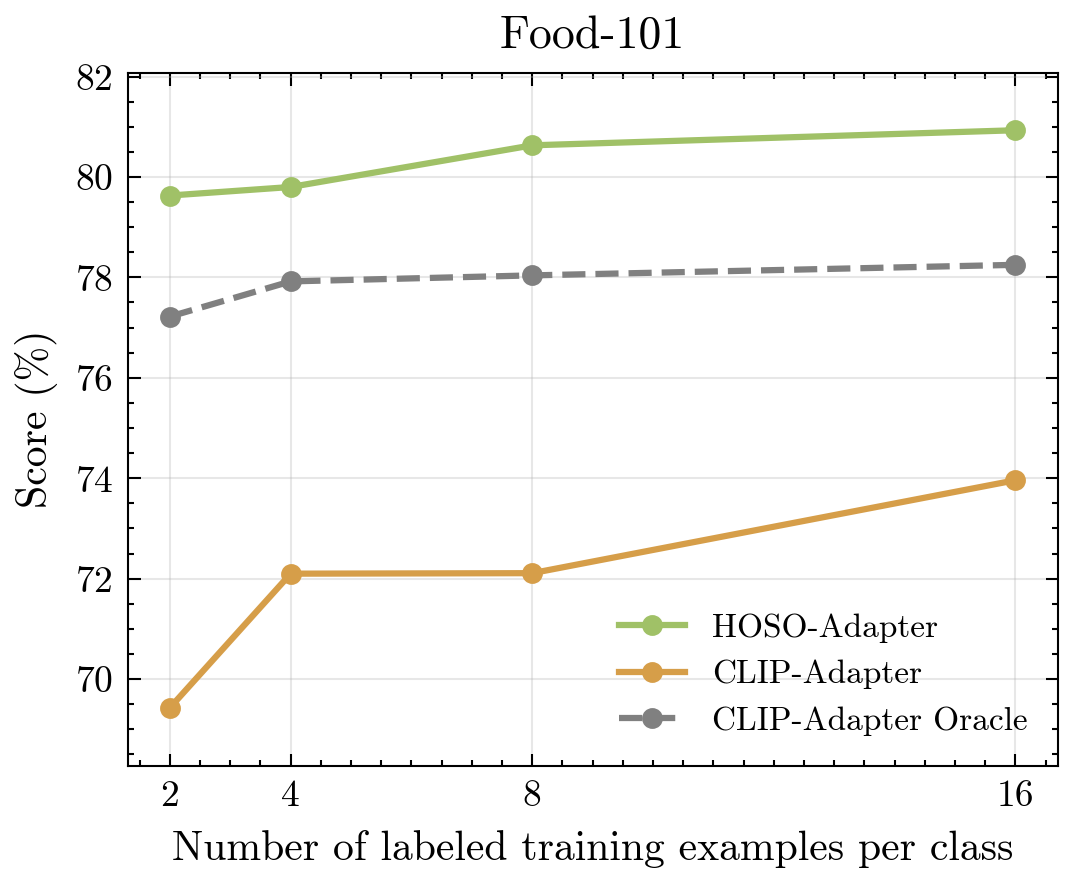}
    \end{minipage}
    \hfill
    \begin{minipage}[t]{0.32\linewidth}
    \centering
    \includegraphics[width=2.2in]{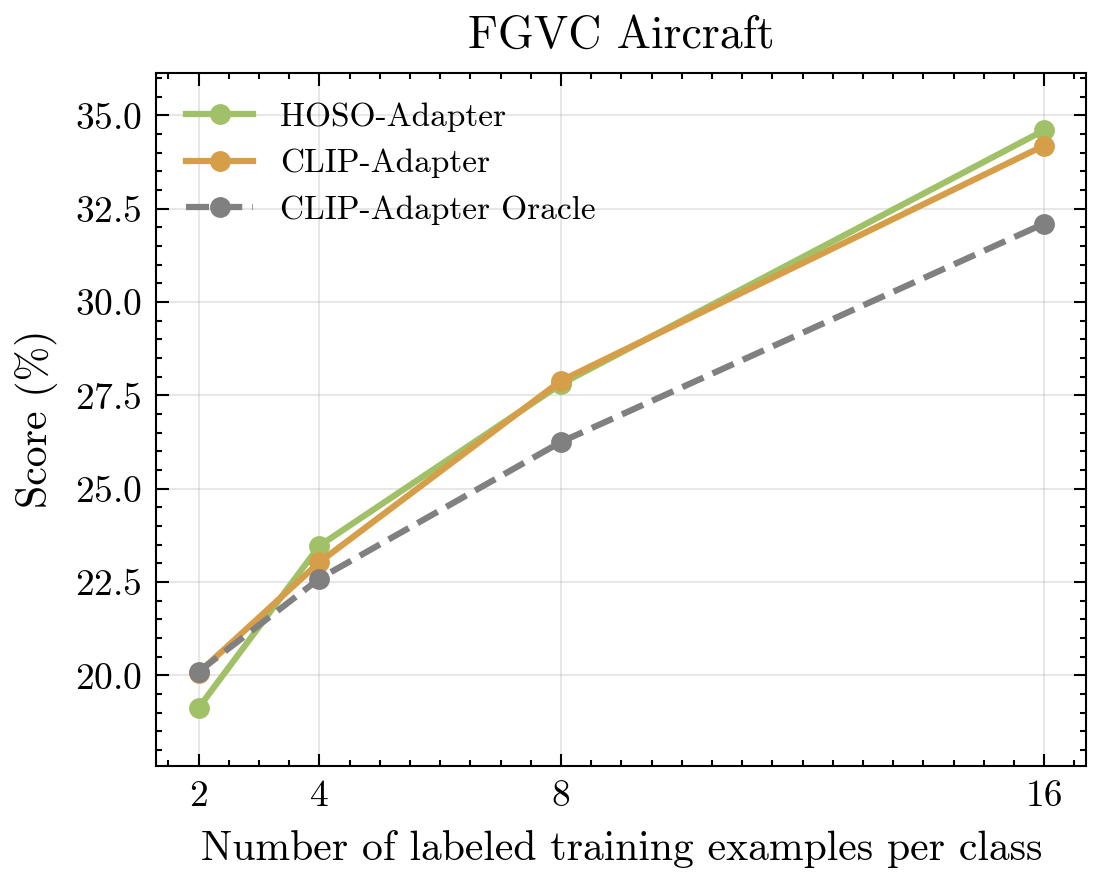}
    \end{minipage}
    \hfill
    \begin{minipage}[t]{0.32\linewidth}
    \centering
    \includegraphics[width=2.2in]{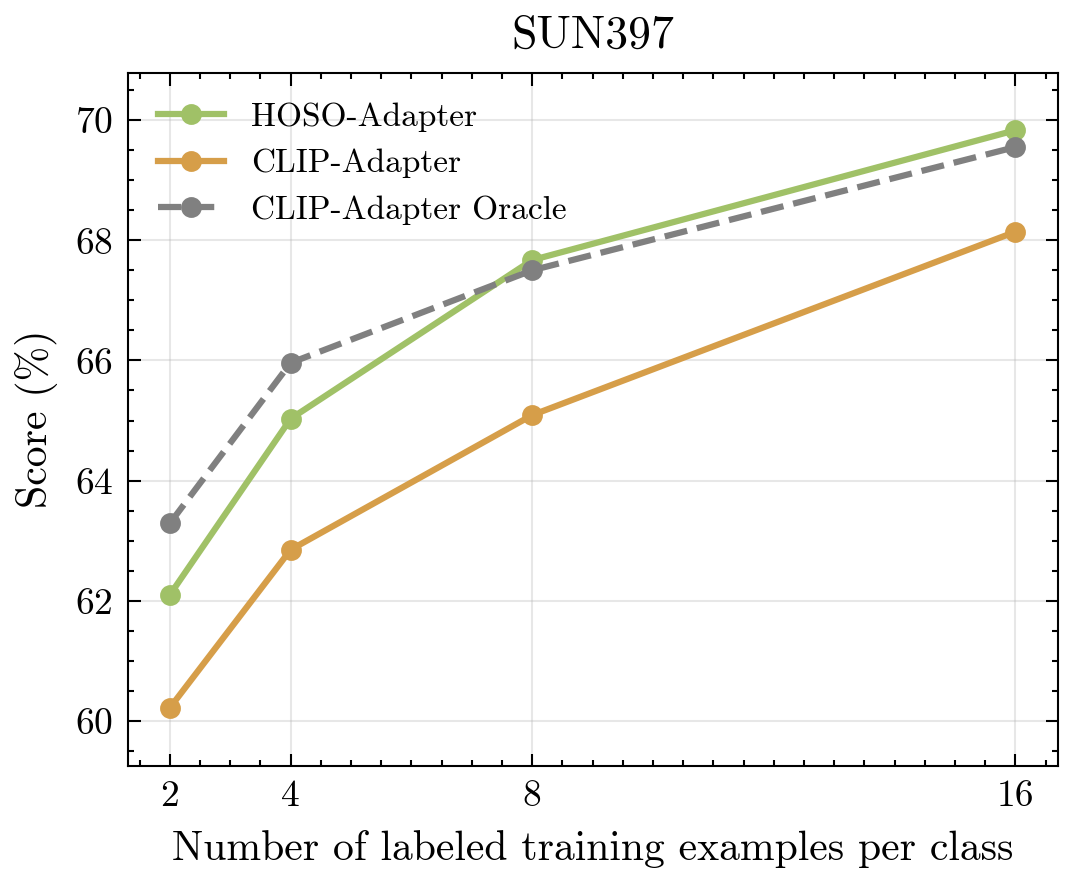}
    \end{minipage}
    
    \vspace{0.05in}
    
    \begin{minipage}[t]{0.32\linewidth}
    \centering
    \includegraphics[width=2.2in]{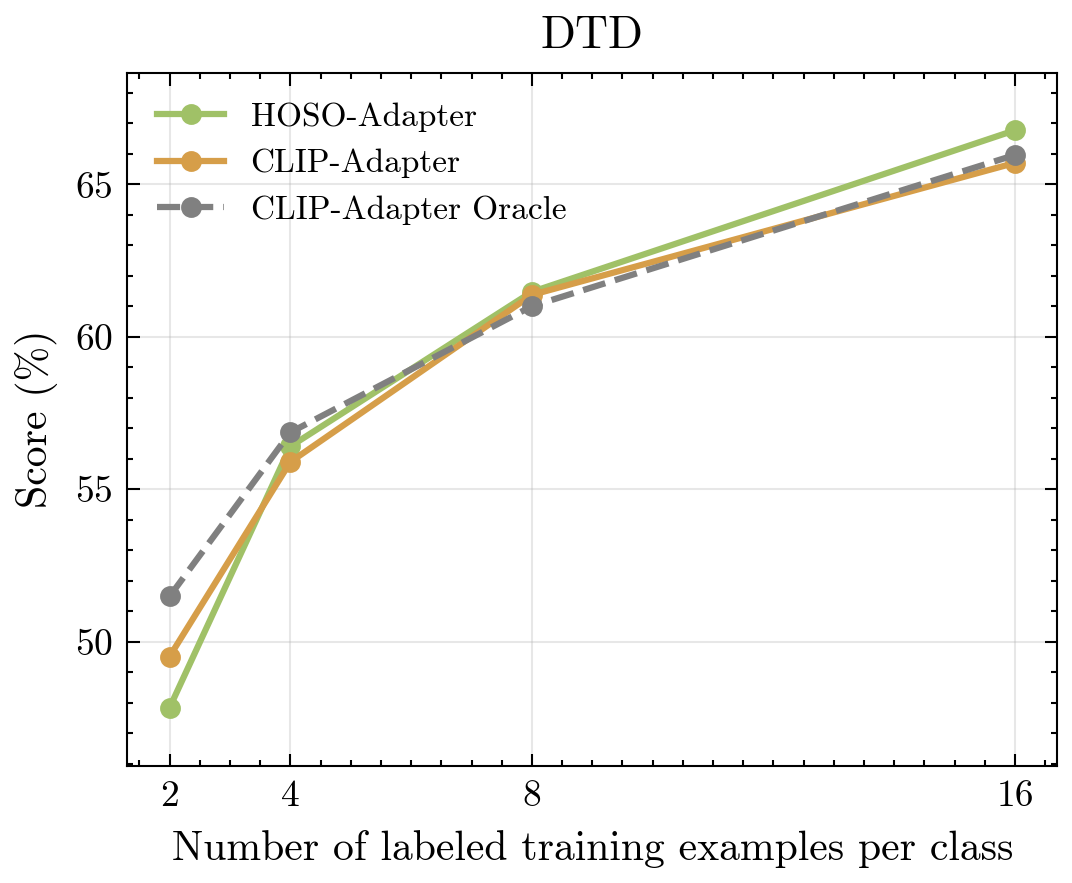}
    \end{minipage}
    \hfill
    \begin{minipage}[t]{0.32\linewidth}
    \centering
    \includegraphics[width=2.2in]{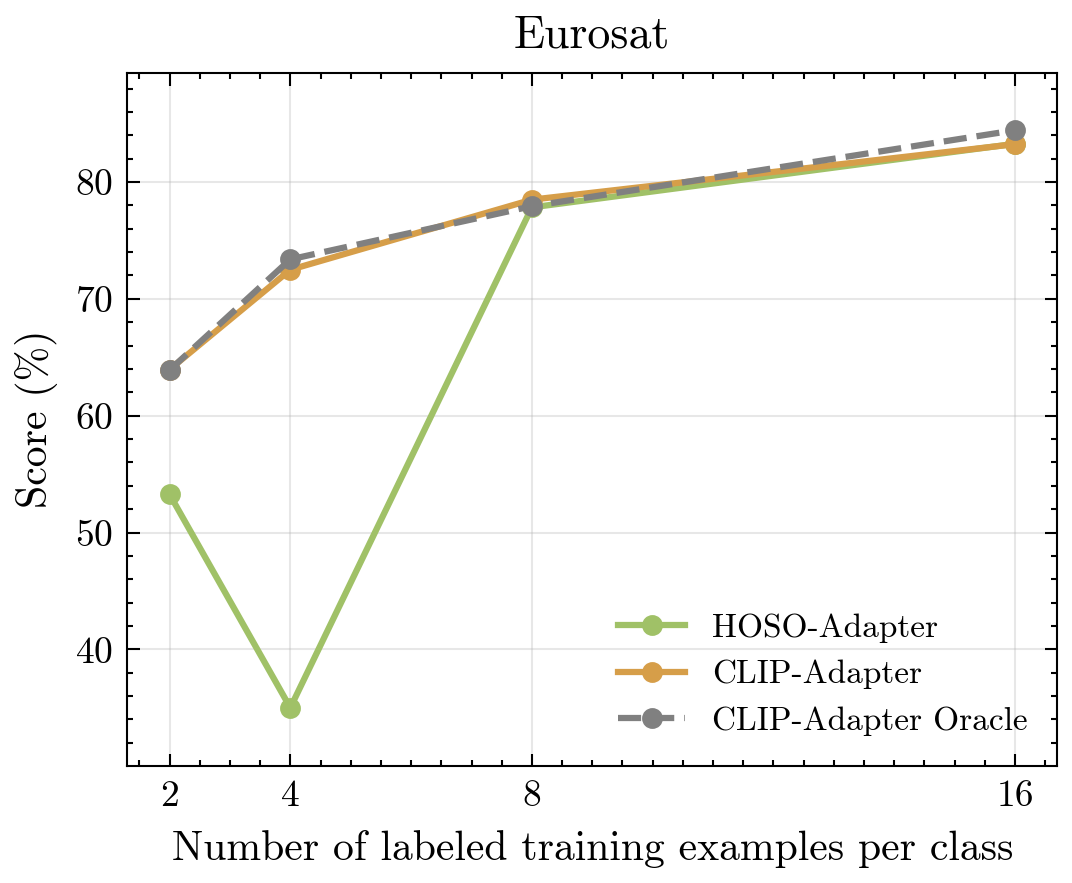}
    \end{minipage}
    \hfill
    \begin{minipage}[t]{0.32\linewidth}
    \centering
    \includegraphics[width=2.2in]{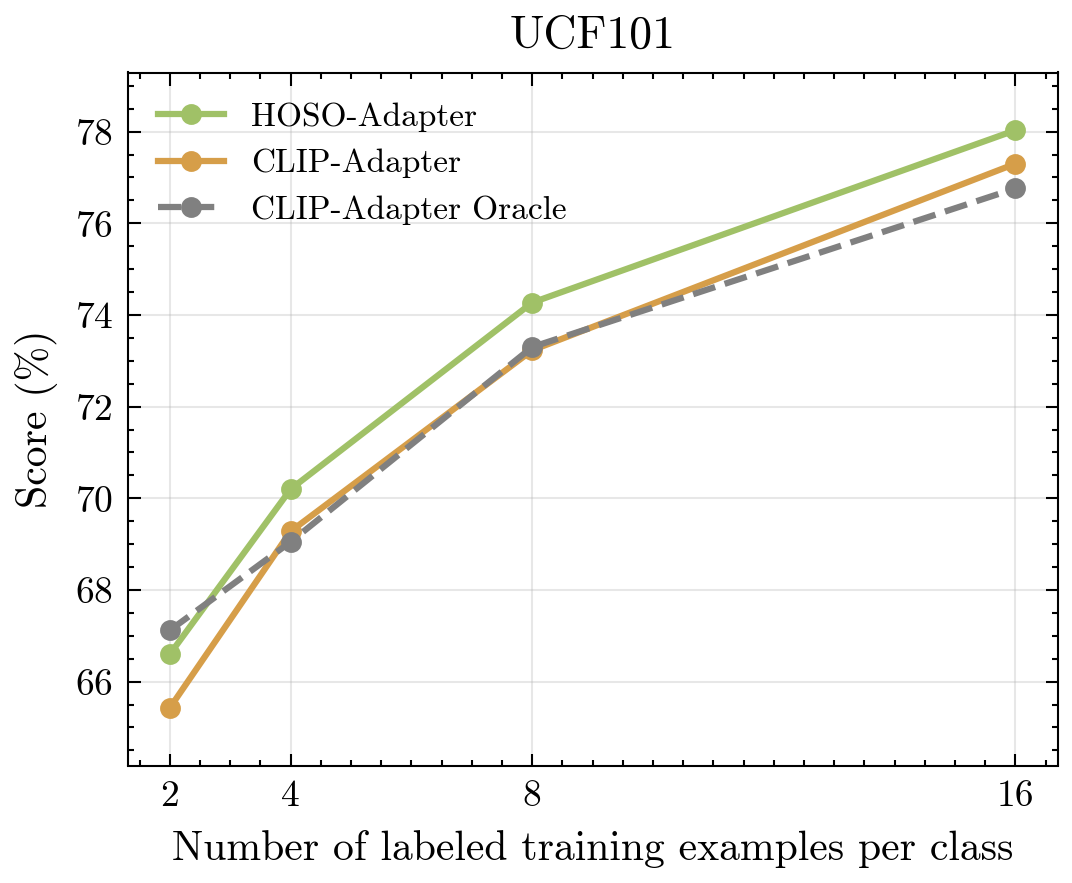}
    \end{minipage}

    \caption{\textbf{HOSO-Adapter performance across few-shot settings} (K=2, 4, 8, 16) on the ResNet-50 backbone with averaged results over 11 datasets. Our method achieves state-of-the-art performance at K=8 and K=16, surpassing even the CLIP-Adapter Oracle baseline. The Oracle baseline, which uses a blending ratio grid-searched on the test set for each dataset, is included for reference and is not directly comparable in the validation-free setting. CLIP-Adapter's results are the validation-free values from \cite{CloserLookFewShot2024_silva-rodriguez}.}
    \label{fig:main_results_per_shot}
    \vspace*{-12pt}
\end{figure*}

\paragraph{Performance across few-shot settings}
A performance comparison across $K=2, 4, 8,$ and $16$ shots is presented in Figure~\ref{fig:main_results_per_shot}. All three methods compared share the same architecture; the only difference is the learning ratio methodology. HOSO-Adapter (green) consistently outperforms the fixed-ratio CLIP-Adapter (yellow), achieving the highest average accuracy at $K=2$, $K=8$, and $K=16$ shots. This advantage is particularly pronounced as the number of shots increases. At K=16, for example, the HOSO-Adapter shows substantial gains over the baseline on both fine-grained datasets like Food101 (+7.0 points) and OxfordPets (+4.5 points), as well as general-purpose benchmarks like ImageNet (+3.9 points). Most notably, at both 8- and 16-shot settings, HOSO-Adapter surpasses the performance of the CLIP-Adapter Oracle (grey), which uses a blending ratio tuned on the test set. 

\subsection{Ablation Study}

Table~\ref{tab:ablation_rn50_16} contains ablation results averaged across ten datasets. The ablations probe three design choices: decoupled optimisation of the adapter and blending ratio, exclusion of the 1-shot cache from adapter training set, and sensitivity to cache size. 

The reference point for the ablation is HOSO-Adapter, which denotes the full method as described in Sec.~\ref{sec:method}. \textit{HOSO-Adapter w/o decoupled training} evaluates the effect of decoupled training of the adapter and the blending ratio; this ablation represents the naive extension of CLIP-Adapter to include a learnt blending ratio into its optimiser. The observed drop to 73.02 demonstrates that simultaneous optimisation harms generalisation because the blending parameter and adapter parameters interact during training. \textit{HOSO-Adapter w/o removing 1-shot from few-shot train set} assesses the importance of holding out the micro-validation set. Here, the 1-shot per class used to tune the blending ratio is retained in the adapter training set rather than excluded. This ablation probes the effect on overfitting. The reduced accuracy (73.35) indicates that excluding the cache reduces overfitting and yields a more reliable estimate of the ratio. \textit{HOSO-Adapter with 2-shots for blending ratio} replaces the 1-shot cache with 2-shots per class (and thus trains the adapter on only 14-shots) to test sensitivity to cache size. Increasing the cache can improve the estimation of the blending weight, but it also reduces the number of training examples available to the adapter. The near-baseline accuracy (76.04) suggests that the benefit of a slightly more optimal blending ratio does not outweigh the reduced adapter performance. \textit{HOSO-Adapter with 8-shots for blending ratio} uses a large cache (8-shots per class) to test the extreme trade-off; allocating half of the examples to the cache reduces the adapter training set by half. The larger drop to 73.68 shows that excessive cache size degrades performance, indicating that a hold-one-shot-out cache, coupled with decoupled optimisation, achieves the best trade-off between blending-ratio learning and sufficient adapter training. 

\begin{table}[t]
\caption{\textbf{Ablation study} on RN50 with 16-shot. We report average accuracy across 10 datasets. Per dataset table provided in Suppl.~Section D.}
\label{tab:ablation_rn50_16}
\vspace{-2mm}
\resizebox{0.47\textwidth}{!}{
\centering
\begin{tabular}{lc}
\toprule
Method & Average\\
\midrule
HOSO-Adapter & \textbf{76.43} \\
HOSO-Adapter w/o decoupled training & 73.02 \\
HOSO-Adapter w/o removing 1-shot from few-shot train set & 73.35 \\
HOSO-Adapter with 2-shots for blending ratio & 76.04 \\
HOSO-Adapter with 8-shots for blending ratio & 73.68 \\

\bottomrule
\end{tabular}
}
\vspace{-5mm}
\end{table}

\begin{figure*}[t]
  \centering
  \includegraphics[width=\textwidth]{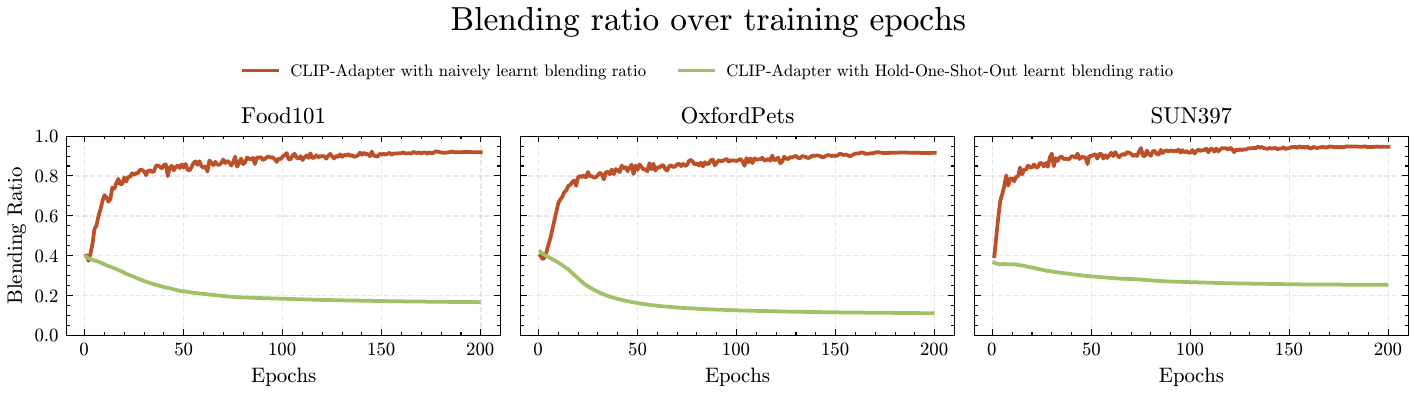}
  \caption{Three example datasets that show the consistent trend: \textit{hold-one-shot-out} reduces the blending ratio (green) compared to the naively learnt blending ratio that enables overfitting on the limited few-shot cases (red). See Suppl.~Section E for full table.}
  \label{fig:learned_ratio_traces_seed1}
\end{figure*}

\begin{figure*}[t]
  \centering
  \includegraphics[width=\textwidth]{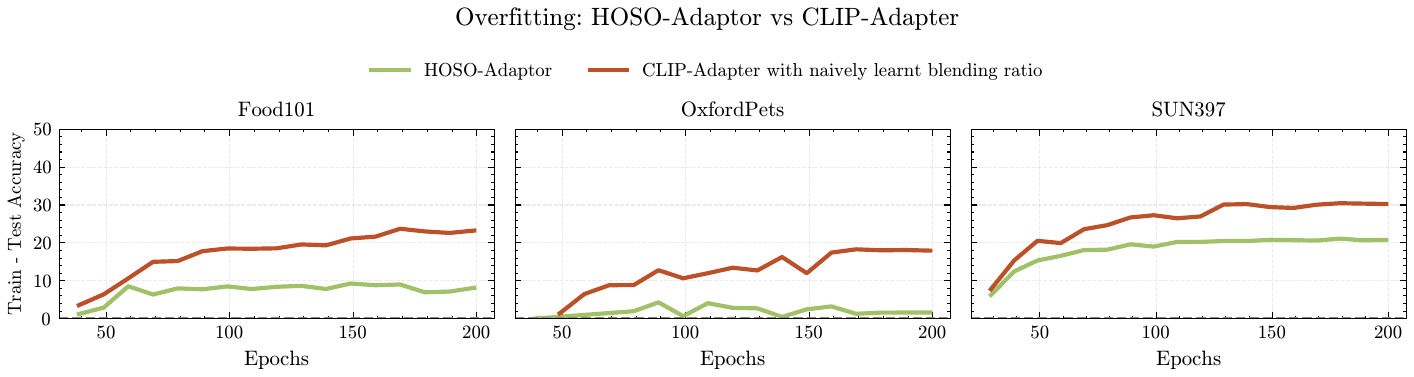}
  \caption{Three example datasets that show the consistent trend: \textit{hold-one-shot-out} (green) overfits less compared to the naively learnt blending ratio (red). See Suppl.~Section F for full table.}
  \label{fig:grid_clip_adapter_compare_selected}
  \vspace{-0.5cm}
\end{figure*}

\section{Discussion}


\paragraph{Method motivation}
The motivation for our hold-one-shot-out method is rooted in the strong empirical correlation between CLIP's 1-shot accuracy and its full-set zero-shot performance (Fig.~\ref{fig:one_per_class_runs_combined}). This correlation suggests that a single example per class is a surprisingly effective proxy for the overall test distribution. HOSO leverages this insight by framing the blending ratio optimisation as a differentiable search for the optimal zero-shot classifier. With the CLIP backbone and text features frozen, the cross-entropy loss on the 1-shot cache directly measures the alignment between the blended image embedding, \(\hat{v}\), and the static text prototypes, \(\{t_c\}\). The gradient with respect to the ratio's logit, \(\alpha_{\text{logit}}\), therefore points in the direction that maximises the likelihood of the correct class under the zero-shot paradigm. Consequently, HOSO performs a gradient-based search for the blending ratio that maximises zero-shot performance on the representative 1-shot samples.  

\paragraph{HOSO as a dynamic regulariser}
The effectiveness of HOSO-Adapter stems from its ability to transform the blending ratio, \(\alpha\), into a dynamic regulariser that prevents overfitting. We analyse the training dynamics of \(\alpha\) in two settings: our decoupled HOSO approach and a naive joint-training baseline (Fig.~\ref{fig:learned_ratio_traces_seed1}).

In the naive setting, where \(\alpha\) and the adapter are optimised on the same few-shot data, \(\alpha\) consistently increases during training. This behaviour prioritises the expressive but over-parameterised adapter, allowing the model to overfit to the limited training examples. In contrast, HOSO's decoupled optimisation on a hold-out set forces \(\alpha\) to assess the adapter's generalisation ability. If the adapter begins to overfit, its features will perform poorly on the hold-out cache, prompting the optimiser to decrease \(\alpha\) and rely more on the robust CLIP prior. This dynamic is observed empirically across all datasets (see Suppl.~Section E), where HOSO consistently learns a more conservative blending ratio. Thus, by decoupling the objectives, HOSO prevents the blending ratio from becoming a catalyst for overfitting and instead serves as a mechanism for preserving generalisation.    

The difference in overfitting behaviour depicted in Fig.~\ref{fig:grid_clip_adapter_compare_selected} validates our observation that HOSO enables the blending ratio to reduce overfitting. For all datasets, the difference between train and test accuracy with the HOSO-Adapter is lower than when the blending ratio is trained without HOSO (see Suppl.~Section F for all datasets, where the same pattern holds).  



\paragraph{Co-adaptation of adapter and blending ratio} A notable finding, as shown in Fig.~\ref{fig:main_results_per_shot}, is that HOSO-Adapter outperforms the grid-searched CLIP-Adapter baseline at higher shot counts (8 and 16). This occurs because a fixed blending ratio, even one that is optimal post hoc, cannot adapt to the adapter's changing state during training.
The adapter's parameters change during training. Initially, its output is noisy; later, it represents task-specific features. A fixed blending ratio, \(\alpha_{\text{fixed}}\), found via grid search may be suboptimal during the early and intermediate training phases. For instance, a high \(\alpha_{\text{fixed}}\) might over-amplify noise from the untrained adapter early on, hindering stable learning.
HOSO-Adapter's learnable ratio, \(\alpha_{\text{logit}}\), adapts throughout training and by continuously evaluating the adapter's generalisation on the hold-out cache, it can dynamically adjust the blend. This allows it to down-weight the adapter's contribution when its features are noisy or begin to overfit, and increase it when they prove effective. 

\section{Conclusion}

HOSO introduces a simple, validation-free mechanism that enables prior-adapter methods to operate in the strict few-shot setting by learning the blending ratio on a 1-shot hold-out cache and training the adapter on the remaining shots. HOSO-Adapter combines this hold-one-shot-out cache with decoupled optimisation, and obtains consistent improvements in few-shot transfer across eleven standard datasets, outperforming CLIP-Adapter even with a dataset-tuned blending ratio on the test set in higher-shot settings. Empirical analysis shows that an HOSO-trained blending ratio prevents the adapter from dominating and overfitting to the small support set.  
\clearpage

{
    \small
    \bibliographystyle{ieeenat_fullname}
    \bibliography{HOSO}
}

\clearpage

\clearpage
\setcounter{page}{1}
\maketitlesupplementary
\setcounter{section}{0}
\renewcommand{\thesection}{\Alph{section}}
\renewcommand{\thefigure}{S\arabic{figure}}
\renewcommand{\thetable}{S\arabic{table}}
\setcounter{figure}{0}
\setcounter{table}{0}

\begin{figure}[t] \centering \includegraphics[width=\columnwidth]{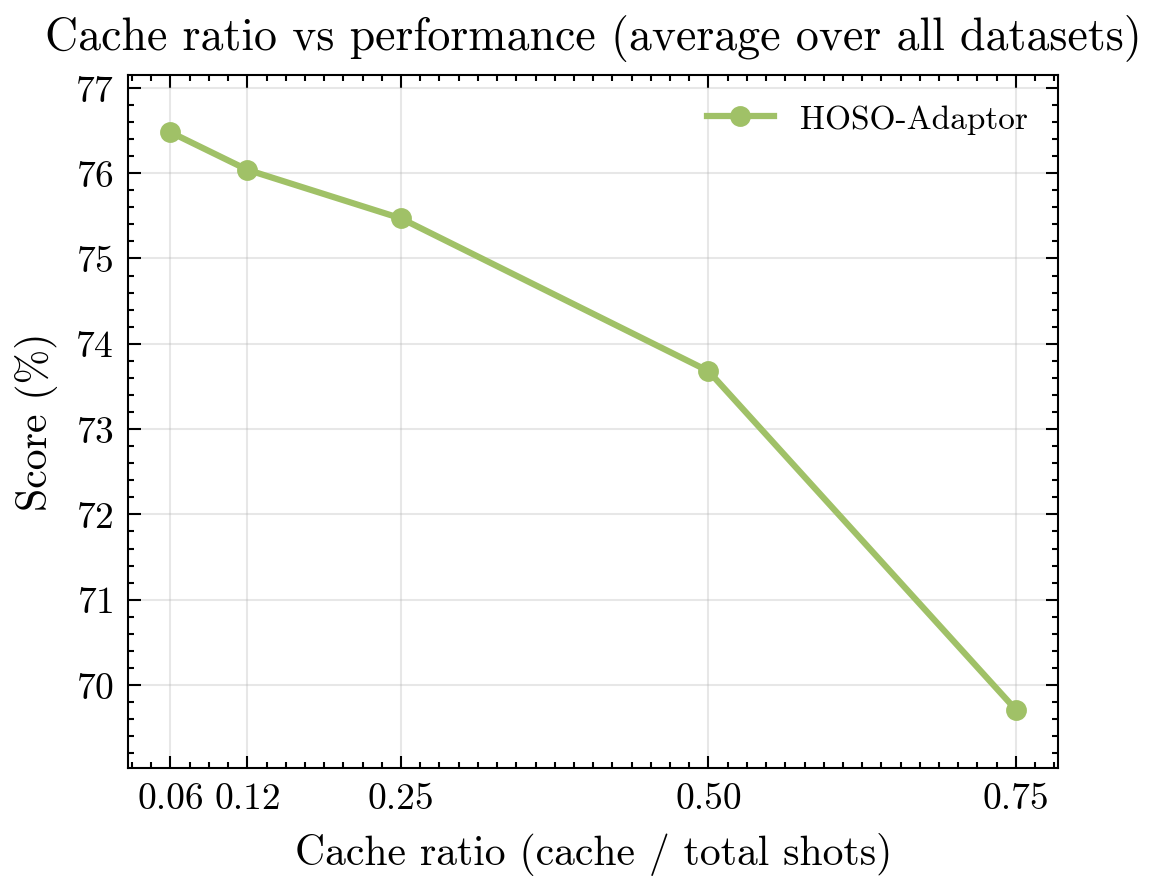} \caption{\textbf{A single hold-out shot is optimal.} An ablation study of the hold-out cache size in the 16-shot setting using ViT-B/16. The average accuracy across ten datasets is plotted against the cache size, expressed as a ratio of the total shots. Performance peaks at a cache size of one (ratio $\approx 0.06$), validating the core design choice. Larger cache sizes lead to performance declines, indicating a trade-off between allocating samples to learn the blending ratio and retaining samples for adapter training.} \label{fig:cache_ratio_ablation} \end{figure}

\section{Related Work Extended}

CLIP few-shot adaptation methods traditionally fall into two categories: prompt learning and adapter-based approaches. Prompt learning optimises global text or visual prompts for downstream tasks, yielding substantial improvements over zero-shot baselines. Prompt learning requires backpropagation through the entire text encoder and access to its weights \cite{LearningPromptVisionLanguage2022_zhoua}. Adapter-based techniques operate in feature space and avoid requiring access to pre-trained model weights because they do not rely on backpropagation. Within adapter-based approaches, training-based variants either train a linear classifier with CLIP features as input or a shallow Multi-Layer Perceptron with a blending ratio. Later works introduce training-free methods that populate a key-value cache using the few-shot examples \cite{TipAdapterTrainingfreeAdaption2022_zhang}. 

Recently, validation-free few-shot has emerged as a distinct setting in which methods must follow a strict few-shot protocol: hyperparameters may not be set using a validation set and must be held constant across datasets \cite{CloserLookFewShot2024_silva-rodriguez}. We adopt this setting because it mirrors real-world scenarios in which the goal is to adapt a large Vision-Language Model (VLM) to a specific domain of interest with limited labelled data. Previous works have used various terms to refer to the blending hyperparameter, including blending ratio, mixing coefficient, weighting parameter, or blending ratio. For clarity and consistency, this work uses the term ``blending ratio'' to refer to the scalar weight that linearly combines the CLIP- and adapted features. 

CLIP-Adapter appends lightweight, learnable bottleneck linear layers to the image branch of CLIP, while keeping the main backbone frozen during few-shot fine-tuning. Adding extra layers can cause overfitting in few-shot settings. To enhance robustness and prevent overfitting, CLIP-Adapter uses a residual connection that integrates new, fine-tuned knowledge with pre-trained representations from the original CLIP backbone. Within this design, the blending ratio \(\alpha\) serves as a critical hyperparameter; CLIP-Adapter performs a comprehensive search over \(\alpha\) for each dataset to select the best-performing configuration. By ablation, Gao \etal \cite{CLIPAdapterBetterVisionLanguage2025_gao} find the optimal \(\alpha\) is 0.6 for the fine-grained Describable Textures Dataset (DTD) and 0.2 for the generic ImageNet. They observe that fine-grained benchmarks favour greater integration of new knowledge. In contrast, generic datasets rely more on existing priors. Setting \(\alpha\) to 0 recovers zero-shot CLIP with no new knowledge, while \(\alpha\) = 1 makes classification depend entirely on the adapted features \cite{CLIPAdapterBetterVisionLanguage2025_gao}. The proposed HOSO-Adapter keeps the architecture and hyperparameters identical to those of CLIP-Adapter. The only difference is that HOSO-Adapter learns the blending ratio, enabling CLIP-Adapter-style methods to compete under the validation-free setting. TipAdapter's prediction function comprises two terms: one adapts and summarises information from the few-shot training set, and the other retains prior knowledge from the original CLIP classifier. The balance between these terms is controlled by a blending ratio \(\alpha\). Zhang \etal find that \(\alpha\) should be set higher when there is a substantial domain gap between the pre-trained and few-shot tasks, since more information from the few-shot set is required in such cases; otherwise, a lower \(\alpha\) suffices \cite{TipAdapterTrainingfreeAdaption2022_zhang}. Proto-Adapter \cite{ProtoAdapterEfficientTrainingFree2024_kato} adapts CLIP via a single-layer class-prototype adapter and linearly blends adapter logits with CLIP zero-shot logits using a blending ratio \(\alpha\). The paper reports per-dataset values of \(\alpha\) without detailing the selection procedure. It is observed that the hardcoded \(\alpha\) values provided in the codebase vary greatly between datasets (\textit{e.g.} Caltech-101 0.8; FGVC-Aircraft 0.2; Oxford 102 Flowers 0.6; and EuroSAT 1.0). Retentive CLIP Adapter Tuning (RCAT) combines CLIP with the temporal processing of a Retentive Network for few-shot video recognition. RCAT employs a specialised adapter-tuning mechanism that modifies the original CLIP architecture to align with the temporal and spatial characteristics of video sequences, thereby improving predictive performance and interpretability. RCAT systematically evaluates the effect of the blending ratio \(\alpha\) on bimodal feature alignment by performing a grid search over \(\alpha\) from 0.1 to 0.9 in increments of 0.1 \cite{RCATRetentiveCLIP2024_xie}. The works listed are a subset of all CLIP adapters that use a grid-searched blending ratio per dataset and thus fall outside the strict few-shot setting. CLIP-Adapter and Proto-Adapter report dataset-dependent optimal values of \(\alpha\): fine-grained benchmarks favour larger \(\alpha\) while generic datasets prefer smaller \(\alpha\); TipAdapter prescribes larger \(\alpha\) when the domain gap is substantial.  

\begin{table*}[t] \caption{\textbf{Main results with RN50 backbone, average over three runs.} Performance of HOSO-Adapter across different few-shot settings.} \label{tab:main_results_rn50} \centering \resizebox{\textwidth}{!}{ \begin{tabular}{lccccccccccccc} \toprule \textbf{Method} & \textbf{Caltech101} & \textbf{DTD} & \textbf{EuroSAT} & \textbf{FGVC} & \textbf{Food101} & \textbf{ImageNet} & \textbf{Flowers} & \textbf{Pets} & \textbf{Cars} & \textbf{SUN397} & \textbf{UCF101} & \textbf{Average} \\ \midrule HOSO-Adapter (2-shot) & 89.10 & 47.83 & 53.27 & 19.13 & 79.63 & 59.63 & 73.90 & 84.10 & 58.00 & 62.10 & 66.60 & 63.03 \\ HOSO-Adapter (4-shot) & 90.17 & 56.40 & 35.00 & 23.47 & 79.80 & 60.70 & 85.50 & 87.73 & 62.43 & 65.03 & 70.20 & 65.13 \\ HOSO-Adapter (8-shot) & 91.40 & 61.47 & 77.83 & 27.80 & 80.63 & 61.75 & 91.90 & 87.73 & 67.13 & 67.67 & 74.27 & 71.78 \\ HOSO-Adapter (16-shot) & 93.03 & 66.77 & 83.27 & 34.60 & 80.93 & 62.93 & 95.07 & 89.47 & 73.80 & 69.83 & 78.03 & 75.25 \\ \bottomrule \end{tabular} } \end{table*}

\begin{table*}[t] \caption{\textbf{Ablation on design choices for HOSO-Adapter (RN50, 16-shot).} The baseline updates the ratio every epoch using a cache of size one per class, removing cached samples from the training set.} \label{tab:ablation_design} \centering \resizebox{\textwidth}{!}{ \begin{tabular}{lccccccccccc} \toprule \textbf{Method} & \textbf{Caltech101} & \textbf{DTD} & \textbf{EuroSAT} & \textbf{FGVC} & \textbf{Food101} & \textbf{Flowers} & \textbf{Pets} & \textbf{Cars} & \textbf{SUN397} & \textbf{UCF101} & \textbf{Average} \\ \midrule Baseline (remove cached samples) & 93.0 & 66.6 & 83.0 & 34.8 & 81.1 & 95.1 & 89.5 & 73.7 & 69.6 & 77.9 & \textbf{76.43} \\ Keep cache in training set & 91.2 & 63.9 & 84.2 & 34.4 & 73.5 & 94.2 & 79.5 & 70.8 & 65.5 & 76.3 & 73.35 \\ Cache size of two per class & 92.6 & 66.5 & 82.1 & 33.9 & 80.8 & 94.6 & 89.0 & 73.6 & 69.5 & 77.8 & 76.04 \\ Cache size of eight per class & 91.7 & 62.6 & 80.9 & 29.3 & 80.7 & 92.5 & 88.0 & 69.1 & 67.3 & 74.7 & 73.68 \\ \bottomrule \end{tabular} } \end{table*}

\begin{table*}[t]
\caption{\textbf{Performance of HOSO-Adapter with data augmentation.} Results use an RN50 backbone.}
\label{tab:rn50_augmentation_results}
\centering
\resizebox{\textwidth}{!}{
\begin{tabular}{lccccccccccr}
\toprule
Shots & Caltech101 & DTD & EuroSAT & FGVCAircraft & Food101 & Flowers102 & OxfordPets & StanfordCars & SUN397 & UCF101 & Average \\
\midrule
2 shots  & 88.5 & 47.7 & 50.7 & 19.1 & 78.8 & 73.6 & 84.7 & 57.5 & 62.0 & 65.7 & 62.83 \\
4 shots  & 89.4 & 56.4 & 32.6 & 22.9 & 79.4 & 85.1 & 87.1 & 61.7 & 65.2 & 69.5 & 64.93 \\
8 shots  & 91.2 & 62.1 & 77.6 & 28.2 & 80.2 & 91.4 & 86.9 & 66.7 & 67.5 & 74.0 & 72.58 \\
16 shots & \textbf{92.5} & \textbf{66.3} & \textbf{81.6} & \textbf{34.9} & \textbf{80.7} & \textbf{94.6} & \textbf{89.2} & \textbf{73.3} & \textbf{69.4} & \textbf{78.0} & \textbf{76.05} \\
\bottomrule
\end{tabular}
}
\end{table*}

\begin{table*}[t] \caption{\textbf{ViT-B/16 16-shot performance comparison.} CLIP-Adapter results are shown with a fixed blending ratio ($\alpha=0.2$) and with the best-performing ratio selected per dataset. HOSO-Adapter results are averaged over three runs.} \label{tab:vitb16_comparison} \centering \resizebox{\textwidth}{!}{ \begin{tabular}{lcccccccccccr} \toprule Method & Caltech101 & DTD & EuroSAT & FGVCAircraft & Food101 & ImageNet & Flowers102 & OxfordPets & StanfordCars & SUN397 & UCF101 & Average \\ \midrule CLIP-Adapter ($\alpha=0.2$) & 94.90 & 59.70 & 70.50 & 34.10 & 89.10 & \textbf{71.50} & 93.10 & 92.60 & 73.90 & 74.20 & 80.40 & 75.82 \\ CLIP-Adapter (best $\alpha$)$^{\dagger}$ & \textbf{95.90} & \textbf{71.70} & \textbf{85.80} & \textbf{45.80} & \textbf{89.30} & \textbf{71.50} & \textbf{97.40} & \textbf{92.70} & \textbf{82.10} & \textbf{75.60} & \textbf{84.00} & \textbf{81.07} \\ HOSO-Adapter (ours) & 95.40 & 70.67 & 85.30 & 43.23 & 88.97 & 70.93 & 97.23 & 92.27 & 81.50 & 74.67 & 83.43 & 80.33 \\ \bottomrule \end{tabular} } \vspace{2mm} \begin{minipage}{\textwidth} \footnotesize $^{\dagger}$ Best blending ratio from the set \{0.2, 0.4, 0.6, 0.7, 0.8\} selected independently per dataset using the test set. \end{minipage} \end{table*}
To the best of our knowledge, only two works have proposed methods to select the blending ratio of CLIP adapters in a validation-free manner, namely SVL-Adapter \cite{SVLAdapterSelfSupervisedAdapter2022_pantazis} and PathCLIP \cite{BridgingPathologyDomain2025_lai}. The SVL-Adapter combines the complementary strengths of vision-language pretraining and self-supervised representation learning and introduces a fully automatic method for selecting the blending hyperparameter \(\alpha\) without requiring held-out, labelled validation data. The method computes \(\alpha\) from CLIP's average prediction confidence on the \(N\) test images of a dataset: \(\alpha = \frac{1}{N}\sum_{i=1}^N \max_k P(y_i = k \mid x_i)\), which directly reflects CLIP's confidence in its zero-shot predictions. This mechanism assumes that when CLIP is not confident, the influence of low-shot learning should increase, automatically adjusting its contribution relative to zero-shot CLIP. The adaptive blending, therefore, leverages the strengths of both zero-shot and low-shot learning to support improved task adaptation in the absence of validation labels \cite{SVLAdapterSelfSupervisedAdapter2022_pantazis}.  

PathCLIP \cite{BridgingPathologyDomain2025_lai} adapts CLIP for pathology and targets learning the blending ratio for its domain of interest. PathCLIP uses Residual Feature Refinement (RFR) as a lightweight adaptation mechanism that tailors CLIP representations to pathology images. As a shallow residual module, RFR is designed to capture salient spatial cues and morphological patterns characteristic of pathological specimens. The self-adaptive blending ratio dynamically balances CLIP knowledge with task-specific features, introducing only a small number of additional trainable parameters. The self-adaptive fusion component, Dual-view Vision Contrastive (DVC), draws inspiration from self-supervised learning approaches such as SimCLR \cite{SimpleFrameworkContrastive2020_chena}. DVC employs two types of augmentations: \textit{weak}, denoting standard flip-and-shift transformations, and \textit{strong}, which consist of more substantial appearance perturbations using methods such as RandAugment \cite{RandAugmentPracticalAutomated2019_cubuk} and CTAugment \cite{REMIXMATCHSEMISUPERVISEDLEARNING_berthelot}. DVC measures the distance between representations obtained from weak and strong augmentations, using cosine similarity. Specifically, the DVC metric calculates the average cosine similarity between encoder outputs for each augmentation across the support set, providing an indicator of the model's consistency under different perturbations. In addition to DVC, support set accuracy serves as a complementary metric for assessing model learning. Since accuracy alone may be insufficient due to the risk of overfitting, particularly when adapting large models to tasks with limited data, PathCLIP combines DVC and accuracy via a bound-constrained mechanism to determine a dynamic blending ratio, $\alpha$. In this work, we use CLIP's average zero-shot prediction confidence (from SVL-Adapter) and PathCLIP's DVC as baselines. Our method differs in that it sets \(\alpha\) as a learnable logit and optimises it independently of the adapter weights via a hold-one-shot-out cache. 

Following \cite{CloserLookFewShot2024_silva-rodriguez}, for the validation-free version of CLIP-Adapter, we set the hyperparameter $\alpha$ to 0.2 for all datasets, as it is the best value reported on ImageNet in the original paper. Similarly, TIP-Adapter sets $\beta$ and $\alpha$ to 1, as recommended in the official repository (https://github.com/gaopengcuhk/Tip-Adapter/issues/13). 

\begin{figure*}[t] \centering \includegraphics[width=\textwidth]{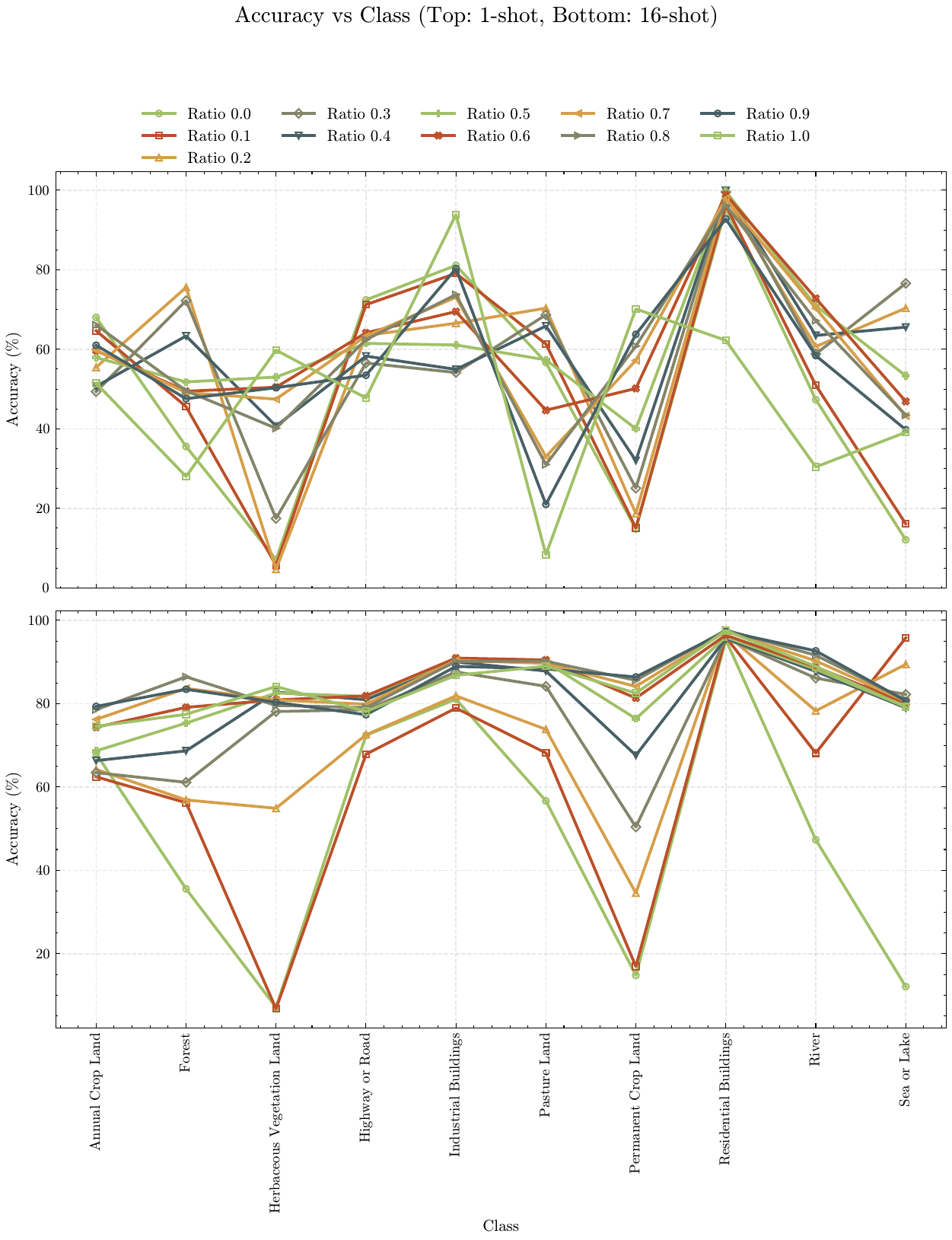} \caption{\textbf{Per-class accuracy as a function of class type for different blending ratios ($\alpha$) on the EuroSAT dataset.} The top plot shows the results for 1-shot learning, while the bottom plot shows 16-shot learning. Each coloured line represents a fixed blending ratio. The analysis highlights that the optimal ratio varies significantly between classes, particularly in the 1-shot scenario.} \label{fig:acc_vs_class} \end{figure*}

\begin{figure*}[t] \centering \includegraphics[width=\textwidth]{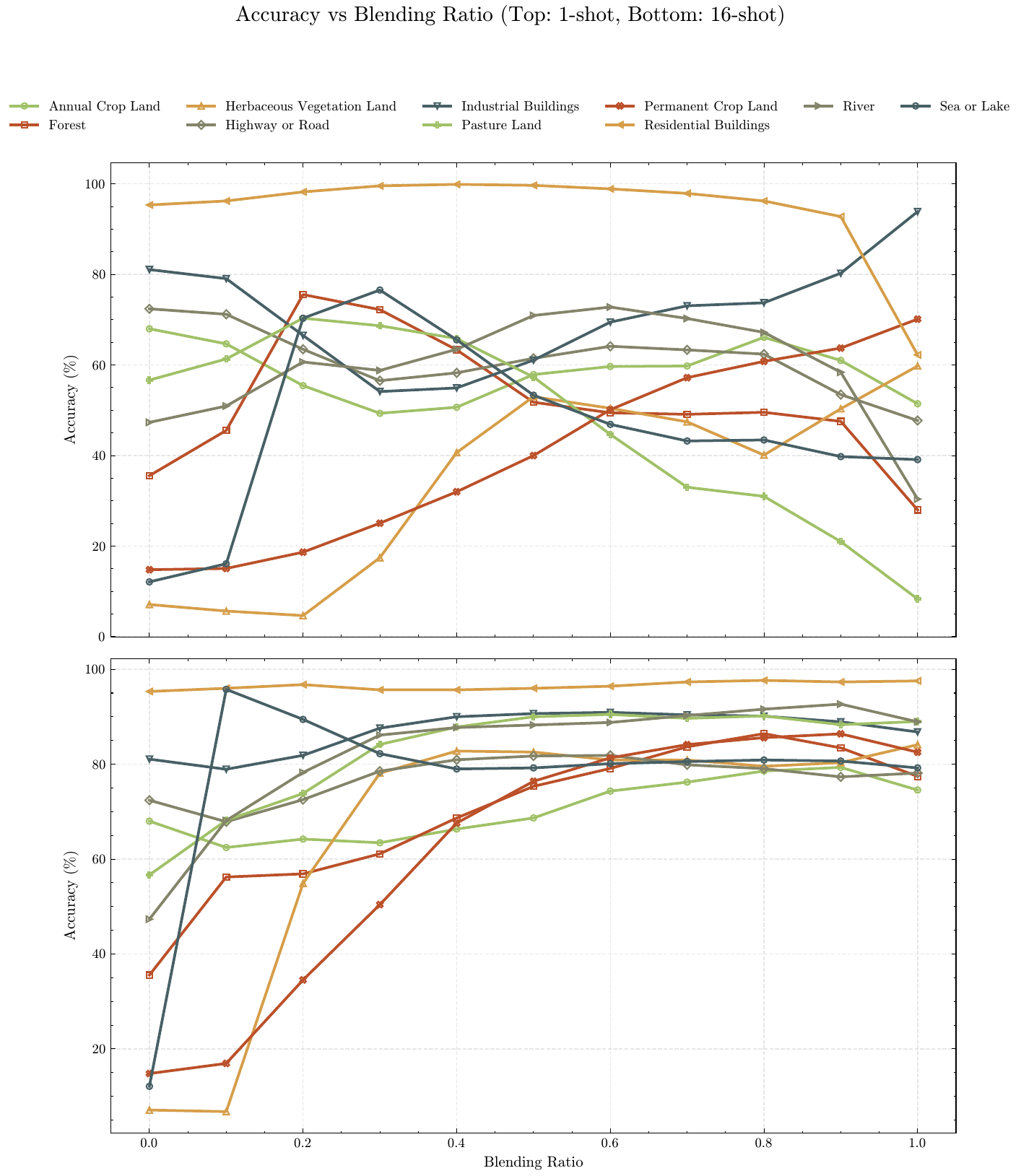} \caption{\textbf{Per-class accuracy as a function of the blending ratio} ($\alpha$) on the EuroSAT dataset. The top plot corresponds to 1-shot and the bottom to 16-shot settings. Each coloured line represents a specific class. The plots show that performance generally improves from the 0-shot baseline ($\alpha=0.0$) to an optimal intermediate ratio, with the 16-shot setting yielding a broader high-performance plateau.} \label{fig:acc_vs_ratio} \end{figure*}

\section{PathCLIP Reimplementation}

The PathCLIP \cite{BridgingPathologyDomain2025_lai} reimplementation employs an online adaptive strategy that adjusts the blending weight $\alpha$ at every training step. Two signals from the current mini-batch guide the update: (1) \textit{Dual-view Vision Contrastive}, which measures the cosine similarity between the CLIP features of weakly and strongly augmented views of the same images, and (2) the mini-batch classification accuracy. A target $\alpha$ is computed as a weighted combination of the DVC score and the inverse of the accuracy. The model's internal $\alpha$ is then updated towards this target value using an exponential moving average, allowing it to dynamically balance the contributions of the CLIP encoder and the adapter throughout training. 

PathCLIP is reimplemented as a CLIP-Adapter variant that learns an adapter and blends it with frozen CLIP image features using an adaptive weight $\alpha$. The implementation loads an official CLIP backbone and freezes both the visual and text encoders. The only trainable component is a lightweight MLP adapter (two linear layers with a bottleneck and ReLU), sized to the CLIP embedding dimensionality (1024 for RN50, 512 for ViT variants). Given an input image, the pipeline computes base CLIP features, passes them through the adapter, and blends in feature space as $\mathrm{image_{fused}} = \alpha \cdot \mathrm{adapter_{out}} + (1 - \alpha) \cdot \mathrm{base_{features}}$. It then normalises image and text features and computes scaled logits using the CLIP logit scale. Similar to CLIP-Adapter, it uses cross-entropy loss and uses dataset-specific single templates as text prompts. The online update of $\alpha$ works as follow: at each training step the procedure computes on GPU, (i) a data variation consistency score as the mean cosine similarity between normalised CLIP features of weak and strong views (weak: horizontal flip plus small translation; strong: RandAugment plus a minimal CTAugment \cite{REMIXMATCHSEMISUPERVISEDLEARNING_berthelot} module), and (ii) the current accuracy. A target $\alpha = w_{\mathrm{dvc}} \cdot \mathrm{dvc} + w_{\mathrm{acc}} \cdot (1 - \mathrm{acc})$ is clamped to $[\alpha_{\mathrm{min}}, \alpha_{\mathrm{max}}]$ and applied with exponential moving average smoothing ($\alpha \leftarrow (1 - \mathrm{smooth}) \cdot \alpha + \mathrm{smooth} \cdot \mathrm{target}$). All augmentation and feature extraction for DVC run on GPU. 

\section{SVL-Adapter Reimplementation}

SVL-Adapter determines the blending weight $\alpha$ data-dependently before training begins. The method first calculates the average zero-shot confidence of the pre-trained CLIP model over all images in the few-shot training set. This average confidence, denoted as $\lambda$, serves as a proxy for zero-shot performance on the target task. The blending weight is set to $\alpha = 1 - \lambda$, thereby giving the adapter more weight when CLIP yields lower zero-shot confidence. The computed $\alpha$ remains fixed for the entire duration of adapter training and subsequent evaluation. We follow the same reimplementation approach as in PathCLIP: freeze CLIP, train only the MLP adapter, and blend the adapter and base features with $\alpha$.


\section{Per-Dataset Ablation Study}

Per-dataset ablations for cache policy and size are reported in Table~\ref{tab:ablation_design}. Updating the ratio at each epoch with a per-class cache of 1 and removing cached samples from training yields the best average (76.43), with gains across most datasets. Retaining cached items in training reduces the average by 3.08 to 73.35, with marked drops on Food101 and Pets; EuroSAT slightly improves but cannot offset broader declines. A cache of two per class is near-neutral (76.04), whilst a cache of eight hurts performance (73.68). A minimal cache with cached samples removed from the training set is the optimal configuration. 

\section{Blending Ratio Over Training}

HOSO-Adapter treats the blending ratio as a dynamic regulariser to curb overfitting. This analysis tracks $\alpha$ under two regimes: the decoupled HOSO approach and a naive joint-training baseline. In the naive case, co-optimising $\alpha$ and the adapter on the same few-shot data steadily increases $\alpha$, overweighting the expressive adapter and leading to overfitting. With HOSO, $\alpha$ is tuned on a hold-out set. It thus reflects generalisation: when the adapter starts to overfit, the features yield lower scores on the hold-out cache, the optimiser lowers $\alpha$ and assigns more weight to the robust CLIP prior. This pattern holds across all datasets except EuroSAT. We posit that this discrepancy arises from the substantial domain gap between CLIP’s pretraining data (natural images) and EuroSAT's satellite images. 

\begin{figure*}[t] \centering \includegraphics[width=\textwidth]{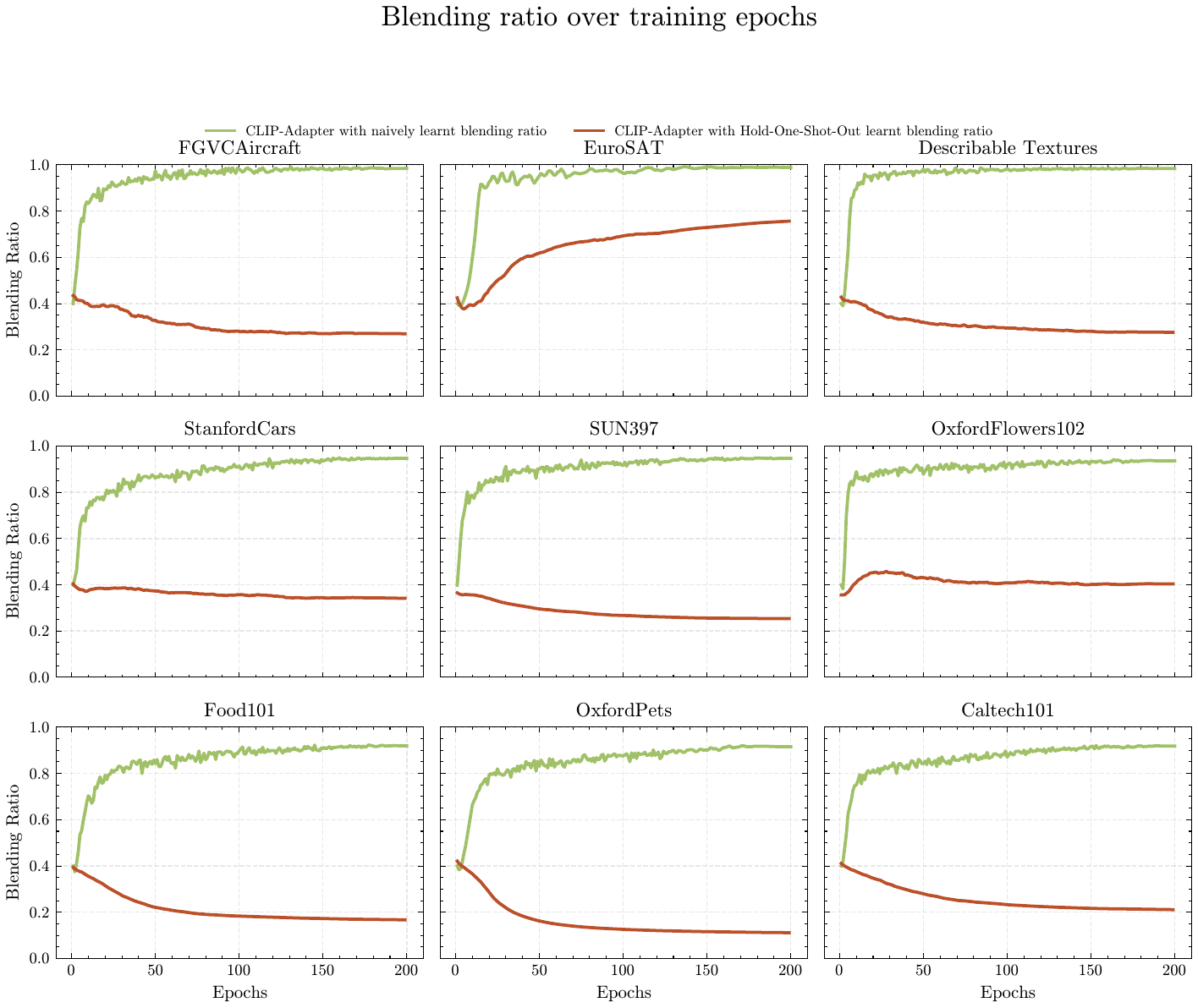} \caption{Most datasets exhibit a trend: \textit{hold-one-shot-out} reduces the blending ratio (green) compared with a learnt blending ratio that leads to overfitting on the limited few-shot cases (red).} \label{fig:learned_ratio_traces} \end{figure*}

\section{Overfitting Analysis}

\begin{figure*}[t] \centering \includegraphics[width=\textwidth]{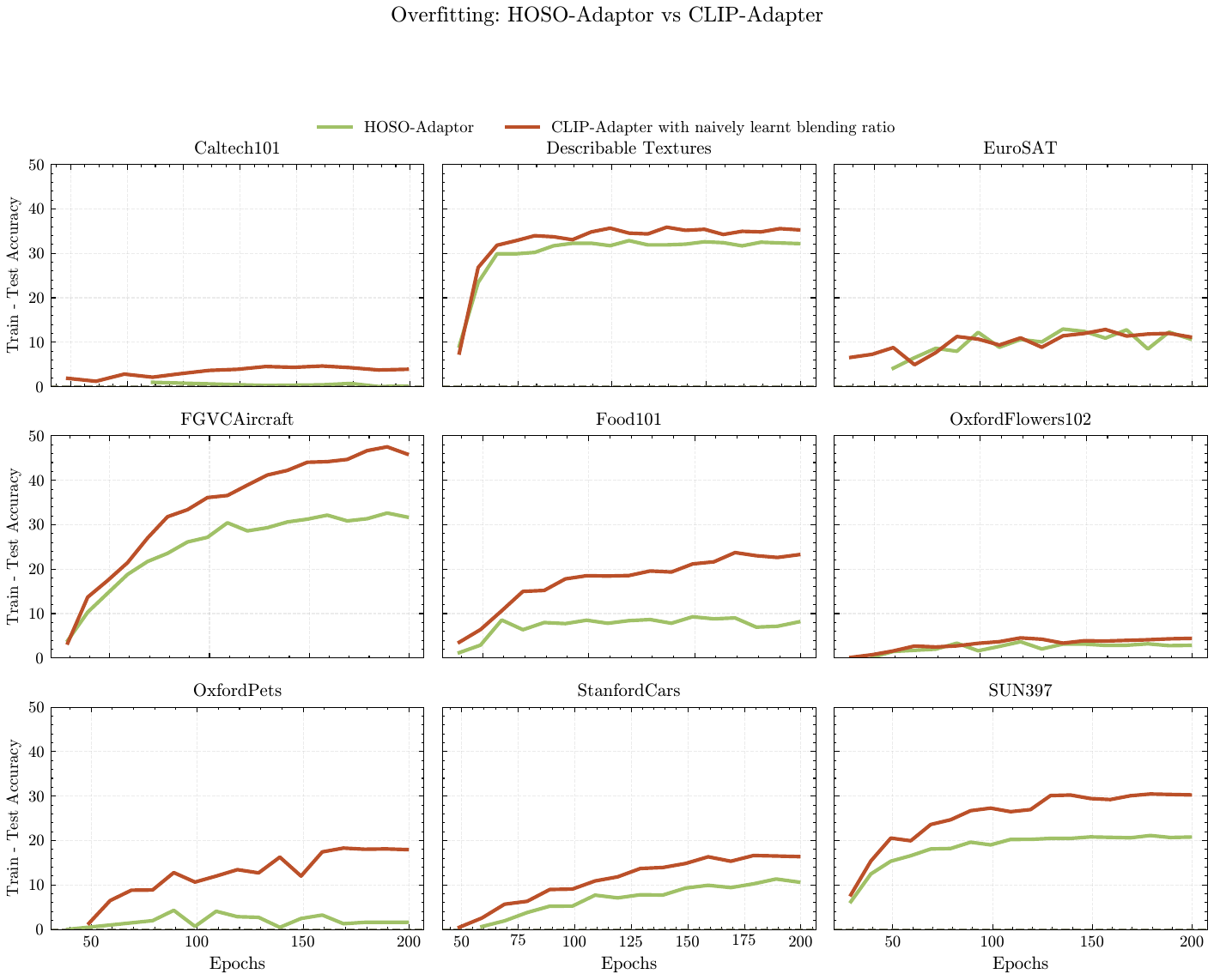} \caption{All datasets exhibit a consistent trend: \textit{hold-one-shot-out} (green) exhibits less overfitting than the naively learnt blending ratio (red).} \label{fig:grid_clip_adapter_compare} \end{figure*}

Figure~\ref{fig:grid_clip_adapter_compare} compares the extent of overfitting for the proposed HOSO-Adaptor with a baseline that directly learns the blending ratio on the full training set. The vertical axis shows the training-test accuracy gap, a direct measure of overfitting, over the full training run. Across the nine benchmark datasets, HOSO-Adaptor (green) consistently maintains a smaller generalisation gap than the baseline (red). The baseline exhibits severe overfitting across several datasets, including Describable Textures, Food101, SUN397, StanfordCars and FGVC Aircraft, where the train-test accuracy gap exceeds 40 percentage points. In contrast, HOSO-Adaptor effectively regularises the model, keeping the gap significantly lower throughout training. This demonstrates the critical role of the decoupled optimisation strategy in reducing overfitting to the few-shot training data and improving generalisation. 

\section{Data Description}

In Table~\ref{tab:datasets_fewshot} we list the 11 datasets and templates used to embed the class labels. These datasets span both fine-grained and coarse-grained domains. We use the same datasets and templates as in CLIP-Adapter.
\begin{table*}[t]
\caption{\textbf{Details of the 11 datasets used for few-shot evaluation.}}
\label{tab:datasets_fewshot}
\centering
\small
\begin{tabular}{lccll}
\toprule
Dataset & Classes & Splits (Tr/V/Te) & Task Category & Prompt Template \\
\midrule
ImageNet        & 1000 & 1.28M / -- / 50k      & objects      & \texttt{"a photo of a \{\}."} \\
Caltech101      & 100  & 4128 / 1649 / 2465    & objects      & \texttt{"a photo of a \{\}."} \\
OxfordPets      & 37   & 2944 / 736 / 3669     & pets         & \texttt{"a photo of a \{\}, a type of pet."} \\
StanfordCars    & 196  & 6509 / 1635 / 8041    & cars         & \texttt{"a photo of a \{\}."} \\
Flowers102      & 102  & 4093 / 1633 / 2463    & flowers      & \texttt{"a photo of a \{\}, a type of flower."} \\
Food101         & 101  & 50.5k / 20.2k / 30.3k & food         & \texttt{"a photo of \{\}, a type of food."} \\
FGVCAircraft    & 100  & 3334 / 3333 / 3333    & aircrafts     & \texttt{"a photo of a \{\}, a type of aircraft."} \\
SUN397          & 397  & 15.9k / 4.0k / 19.9k  & scenes       & \texttt{"a photo of a \{\}."} \\
DTD             & 47   & 2820 / 1128 / 1692    & textures     & \texttt{"\{\} texture."} \\
EuroSAT         & 10   & 13.5k / 5.4k / 8.1k   & satellite    & \texttt{"a centered satellite photo of \{\}."} \\
UCF101          & 101  & 7639 / 1898 / 3783    & actions      & \texttt{"a photo of a person doing \{\}."} \\
\bottomrule
\end{tabular}
\end{table*}


\section{Blending Ratio Grid Search}

Increasing the blending weight on the adapter with a ViT-B/16 backbone in the 16-shot setting has variable effects. Average accuracy rises from 74.82 at $\alpha=0.2$ to a peak of 80.62 at $\alpha=0.6$, then slightly decreases at $\alpha=0.8$. Fine-grained and distribution-shifted datasets (DTD, EuroSAT, FGVC Aircraft, Stanford Cars) benefit most from higher adapter contribution, while tasks with strong CLIP priors or broader scene cues (OxfordPets, SUN397, UCF101) favour lower to mid weights (around $\alpha=0.4$). Caltech101 and Flowers102 remain stable across different values of $\alpha$ in the 0.4 - 0.8 range. 

\begin{table}[t] \caption{\textbf{CLIP-Adapter ViT-B/16 blending search (16-shot).} Classification accuracy for different blending weights ($\alpha$). A low $\alpha$ favours CLIP features, while a high $\alpha$ favours the adapter.} \label{tab:vitb16_residual_search} \centering \begin{tabular}{lcccc} \toprule \textbf{Dataset} & \textbf{$\alpha=0.2$} & \textbf{$\alpha=0.4$} & \textbf{$\alpha=0.6$} & \textbf{$\alpha=0.8$} \\ \midrule Caltech101 & 94.9 & \textbf{95.7} & \textbf{95.7} & 95.3 \\ DTD & 59.7 & 69.7 & 71.5 & \textbf{71.6} \\ EuroSAT & 70.5 & 80.3 & 84.1 & \textbf{85.8} \\ FGVCAircraft & 34.1 & 40.9 & 44.1 & \textbf{45.8} \\ Flowers102 & 93.1 & 97.0 & \textbf{97.3} & 96.8 \\ OxfordPets & 92.6 & \textbf{92.7} & 92.2 & 90.7 \\ StanfordCars & 73.9 & 79.7 & \textbf{81.9} & 81.7 \\ SUN397 & 74.2 & \textbf{75.6} & 75.1 & 73.7 \\ UCF101 & 80.4 & \textbf{84.0} & 83.7 & 83.5 \\ \midrule \textbf{Average} & 74.82 & 79.51 & \textbf{80.62} & 80.54 \\ \bottomrule \end{tabular} \end{table}

Following the ViT-B/16 sweep, we repeat the blending-weight grid search for the RN50 backbone under the same 16-shot protocol. We interpolate between the frozen CLIP features and the adapter outputs using $\alpha \in \{0.2, 0.4, 0.6, 0.8\}$, where a higher $\alpha$ increases the adapter contribution. Compared to ViT-B/16, RN50's weaker visual prior makes the optimal $\alpha$ more dataset-dependent: larger blending ratios often help on distribution-shifted or fine-grained tasks (e.g., EuroSAT, FGVC Aircraft, Stanford Cars), while smaller ratios suffice when CLIP priors are strong (e.g., Food101, Pets). Table~\ref{tab:vitb16_residual_search} reports the resulting per-dataset accuracies. 

\begin{table}[t]
\caption{\textbf{CLIP-Adapter RN50 blending ratio search (16-shot).}}
\label{tab:rn50_residual_search}
\centering
\begin{tabular}{lcccc}
\toprule
\textbf{Dataset} & \textbf{$\alpha=0.2$} & \textbf{$\alpha=0.4$} & \textbf{$\alpha=0.6$} & \textbf{$\alpha=0.8$} \\
\midrule
Caltech101     & 91.4          & \textbf{92.7} & 92.6          & 91.8   \\
DTD            & 59.2          & 66.1          & \textbf{67.0} & 66.2 \\
EuroSAT        & 65.4          & 81.4          & 83.3          & \textbf{84.0} \\
FGVCAircraft   & 24.5          & 32.1          & 35.0          & \textbf{35.6} \\
Food101        & \textbf{81.3} & 79.9          & 77.9          & 75.7   \\
Flowers102  & 89.7          & 94.0          & 94.7 & \textbf{94.8}   \\
OxfordPets     & \textbf{88.6} & 87.9          & 85.1          & 83.9   \\
StanfordCars   & 67.6          & 73.5          & \textbf{74.7} & 73.2 \\
SUN397         & 69.8          & \textbf{70.3} & 69.4          & 68.0   \\
UCF101         & 76.2          & 78.8          & \textbf{78.9} & 77.9 \\
\bottomrule
\end{tabular}
\end{table}

\section{Design Choices Ablation}

Table \ref{tab:residual_comparison_vit} shows that learning the blending weight yields the highest average accuracy, slightly improves on a fixed blending weight, and clearly outperforms random sampling. HOSO-Adapter achieves the best mean accuracy, while CLIP-Adapter with a fixed blending weight leads on some datasets. The fixed-value CLIP-Adapter (0.6) is shown as an upper bound for fixed-ratio performance, as it is selected based on the best test-set performance across all datasets via grid search. This ablation demonstrates that learning the blending ratio as the default design choice yields clear improvement over random sampling and even outperforms the fixed-ratio upper bound in the 16-shot setting. 

\begin{table*}[t] \caption{\textbf{Comparison of different blending ratio strategies for CLIP-Adapter (ViT-B/16, 16-shot).} HOSO-Adapter uses a dynamically learned blending ratio.} \label{tab:residual_comparison_vit} \centering \begin{tabular}{lccc} \toprule \textbf{Dataset} & \textbf{Fixed $\alpha=0.6$} & \textbf{Random $\alpha \in [0,1]$} & \textbf{HOSO-Adapter} \\ \midrule Caltech101 & 95.7 & 95.4 & \textbf{96.1} \\ DTD & 71.5 & 67.0 & \textbf{72.5} \\ EuroSAT & 84.1 & 79.6 & \textbf{86.6} \\ FGVCAircraft & 44.1 & 37.3 & \textbf{46.4} \\ Flowers102 & \textbf{97.3} & 94.0 & 97.1 \\ OxfordPets & \textbf{92.2} & 88.3 & 90.4 \\ StanfordCars & \textbf{81.9} & 69.9 & 80.4 \\ SUN397 & \textbf{75.1} & 68.2 & 74.4 \\ UCF101 & \textbf{83.7} & 79.4 & 83.3 \\ \midrule \textbf{Average} & 80.62 & 75.46 & \textbf{80.80} \\ \bottomrule \end{tabular} \end{table*}

\begin{table}[t]
\caption{\textbf{CLIP-Adapter ViT-B/16 feature vs. logit blending (16-shot; top-1 accuracy, \%).} Both blending methods use a blending weight of $\alpha=0.7$.}
\label{tab:vitb16_feature_vs_logit}
\centering
\begin{tabular}{lcc}
\toprule
Dataset & Feature Blend & Logit Blend \\
\midrule
Caltech101 & \textbf{95.9} & 94.3 \\
DTD & \textbf{71.7} & 70.7 \\
EuroSAT & 85.1 & \textbf{85.5} \\
FGVC Aircraft & 45.4 & \textbf{46.4} \\
Flowers102 & \textbf{97.4} & 96.9 \\
OxfordPets & \textbf{91.6} & 90.5 \\
Stanford Cars & 82.1 & \textbf{82.4} \\
SUN397 & \textbf{74.4} & 72.5 \\
UCF101 & \textbf{83.7} & 82.5 \\
\midrule
Average & \textbf{80.8} & 80.2 \\
\bottomrule
\end{tabular}
\end{table}



There are two strategies for combining knowledge from the pre-trained CLIP model and a trained adapter. By experimentation, we determine whether it is more effective to blend information in the feature space or in the final logit space. The comparison in Table~\ref{tab:vitb16_feature_vs_logit} uses a fixed blending ratio weight of $\alpha=0.7$. The two methods are as follows: Feature Blending, which performs a linear interpolation between the feature vector from the original CLIP image encoder, $\mathbf{f}_{\text{CLIP}}$, and the output of the adapter, $\mathbf{f}_{\text{adapter}}$. The approach then normalises the resulting blended feature vector, $\mathbf{f}_{\text{blend}} = \alpha \cdot \mathbf{f}_{\text{adapter}} + (1 - \alpha) \cdot \mathbf{f}_{\text{CLIP}}$, and computes the final logits. This method integrates the adapter's adjustments directly into the image representation before the classification head. The second approach, Logit Blending, operates at the end of the pipeline. It first computes two sets of logits: one using the original CLIP features ($\mathbf{l}_{\text{CLIP}}$) and another using the adapter's features ($\mathbf{l}_{\text{adapter}}$). The final prediction is a weighted average of these two logit vectors: $\mathbf{l}_{\text{final}} = \alpha \cdot \mathbf{l}_{\text{adapter}} + (1 - \alpha) \cdot \mathbf{l}_{\text{CLIP}}$. This constitutes a form of model ensembling at the prediction level. As shown in Table~\ref{tab:vitb16_feature_vs_logit}, neither method is universally superior. However, feature blending shows a slight advantage on several datasets, which motivates its use in the primary experiments. 

Table~\ref{tab:ablation_lr} indicates that the blending ratio optimiser exhibits relatively low sensitivity to the learning rate in the 16-shot RN50 setting, with a stable plateau from 0.001 to 0.4. The average accuracy peaks at 0.1, matching the per-dataset optimum on DTD, EuroSAT and UCF101. These results motivate choosing 0.1 as the learning rate for the blending ratio. 

\begin{table}[t]
\caption{\textbf{Ablation on the learning rate for the blending ratio optimiser.} Experiments use RN50 with 16-shots.}
\label{tab:ablation_lr}
\centering
\begin{tabular}{lcccc}
\toprule
\textbf{Dataset} & \textbf{0.001} & \textbf{0.01} & \textbf{0.1} & \textbf{0.4} \\
\midrule
DTD & 65.5 & 66.5 & \textbf{66.7} & 66.6 \\
EuroSAT & 83.9 & 83.6 & \textbf{84.0} & 83.0 \\
UCF101 & 76.2 & 77.4 & \textbf{78.1} & 77.9 \\
\midrule
\textbf{Average} & 75.20 & 75.83 & \textbf{76.27} & 75.83 \\
\bottomrule
\end{tabular}
\end{table}


Figure \ref{fig:cache_ratio_ablation} shows that HOSO-Adapter achieves peak average accuracy when the hold-out cache uses one shot, approximately six per cent of the sixteen-shot budget. Increasing the cache beyond one shot consistently reduces accuracy across ten datasets, indicating a trade-off between reliable blending-ratio estimation and sufficient adapter training data. We therefore propose a hold-one-shot-out cache. 

\section{Detailed Results}

\begin{table*}[t] \caption{\textbf{ImageNet accuracy with RN50 backbone.} Average of three runs.} \label{tab:imagenet_rn50} \centering \begin{tabular}{lccccc} \toprule \textbf{Method} & \textbf{2-shot} & \textbf{4-shot} & \textbf{8-shot} & \textbf{16-shot} & \textbf{Average} \\ \midrule CLIP-Adapter & 59.63 & 60.67 & 61.67 & 63.07 & 61.26 \\ HOSO-Adapter & 59.63 & 60.70 & 61.75 & 62.93 & 61.25 \\ \bottomrule \end{tabular} \end{table*}

Table~\ref{tab:imagenet_rn50} shows that HOSO-Adapter matches CLIP-Adapter on ImageNet with an RN50 backbone across few-shot settings. Both methods improve as the number of shots increases, and their averages are effectively identical (61.25 versus 61.26). However, CLIP-Adapter in this case uses the test set to grid-search the optimal blending ratio for each dataset. In contrast, HOSO-Adapter learns the blending ratio under the strict few-shot protocol. 

Table \ref{tab:main_results_rn50} shows that HOSO-Adapter with an RN50 backbone improves steadily with more shots, raising the average accuracy from 63.03 at 2-shot to 75.25 at 16-shot. The largest gains occur for EuroSAT, Flowers102, DTD and Stanford Cars, with improvements of +30.0, +21.17, +18.94 and +15.80 percentage points, respectively, while Food101 and ImageNet change modestly. Performance on FGVC Aircrafts remains comparatively low but increases with shot count. 

Table \ref{tab:main_results_vitb16} shows that HOSO-Adapter scales reliably with additional supervision: the mean accuracy improves from 69.76 at 2-shot to 80.33 at 16-shot. The most significant gains occur on EuroSAT (+26.07) and DTD (+20.27). Consistent improvements are observed on fine-grained datasets such as FGVC Aircraft and Stanford Cars. Performance on datasets already aligned with CLIP, including Food101 and Caltech101, remains high with only marginal changes. EuroSAT displays a non-monotonic outcome, with 4-shot underperforming 2-shot. Inspection of the logs and code does not reveal an apparent reason for the decrease with the 4-shot setting. As this occurs only for this dataset and the same lower results are observed over three randomised seed runs, the outcome likely reflects variability in performance due to the significant domain gap between satellite images and CLIP's pretraining images. 

\begin{table*}[t]
\caption{\textbf{Main results with ViT-B/16 backbone, averaged over three runs.} Performance of HOSO-Adapter across different few-shot settings.}
\label{tab:main_results_vitb16}
\centering
\resizebox{\textwidth}{!}{
\begin{tabular}{lcccccccccccc}
\toprule
\textbf{Method} & \textbf{Caltech101} & \textbf{DTD} & \textbf{EuroSAT} & \textbf{FGVC} & \textbf{Food101} & \textbf{ImageNet} & \textbf{Flowers} & \textbf{Pets} & \textbf{Cars} & \textbf{SUN397} & \textbf{UCF101} & \textbf{Average} \\
\midrule
HOSO-Adapter (2-shot)  & 94.20 & 50.40 & 59.23 & 27.73 & 88.20 & 67.53 & 82.07 & 90.00 & 67.73 & 67.03 & 73.27 & 69.76 \\
HOSO-Adapter (4-shot)  & 94.67 & 59.57 & 45.10 & 32.57 & 87.97 & 68.67 & 89.43 & 91.63 & 70.90 & 70.13 & 77.10 & 71.61 \\
HOSO-Adapter (8-shot)  & 94.83 & 65.40 & 79.87 & 37.47 & 88.90 & 69.83 & 95.17 & 91.57 & 75.37 & 72.60 & 80.73 & 77.43 \\
HOSO-Adapter (16-shot) & \textbf{95.40} & \textbf{70.67} & \textbf{85.30} & \textbf{43.23} & \textbf{88.97} & \textbf{70.93} & \textbf{97.23} & \textbf{92.27} & \textbf{81.50} & \textbf{74.67} & \textbf{83.43} & \textbf{80.33} \\
\bottomrule
\end{tabular}
}
\end{table*}


Table~\ref{tab:vitb16_comparison} shows that HOSO-Adapter achieves an average of 80.33 at 16-shot, outperforming CLIP-Adapter (validation-free, hence $\alpha=0.2$) by more than four percentage points while trailing the per-dataset-tuned CLIP-Adapter by less than one point. Together with the trend in Table~\ref{tab:main_results_vitb16}, these results indicate that HOSO-Adapter delivers strong 16-shot performance without per-dataset hyperparameter search on the test set. 

\section{Class-level Perspective}

Figure~\ref{fig:acc_vs_class} illustrates the classification accuracy for each class within the EuroSAT dataset, comparing performance across different blending ratios ($\alpha$). A ratio of zero corresponds to the zero-shot CLIP model, while a ratio of one relies solely on features from the trained adapter. The top panel (1-shot) shows significant performance variability, indicating that the optimal blending ratio is highly class-dependent. For instance, accuracy for classes such as ``Residential Buildings'' is high with ratios ranging from 0.7 to 0.9, whereas for ``Herbaceous Vegetation Land'' the best performance is around 0.8, despite a very low zero-shot baseline. This indicates that there is no single optimal ratio for all classes within a dataset, especially in low-data regimes. The bottom panel (16-shot) shows a marked improvement in overall accuracy and reduced sensitivity to the exact ratio. With more shots, the adapter becomes more robust, and most ratios above 0.4 yield strong, comparable results that consistently outperform the zero-shot baseline. 

Figure~\ref{fig:acc_vs_ratio} presents a complementary view, plotting the accuracy of each class as a function of the blending ratio, $\alpha$. This visualisation clarifies the impact of increasing the adapter's influence. In the 1-shot scenario (top), most classes exhibit a unimodal performance curve, starting at the zero-shot baseline ($\alpha=0$), peaking at an intermediate ratio, and often declining as $\alpha$ approaches one. This pattern suggests that while the adapter provides crucial task-specific information, completely discarding the general-purpose features from the pre-trained CLIP encoder is detrimental. The optimal peak is class-specific, with some classes like ``Forrest'' peaking at lower adapter influence (\eg, $\alpha \approx 0.2$) and others like ``Permanent Crop Land'' peaking at higher influence (\eg, $\alpha \approx 1$). In the 16-shot setting (bottom), the performance curves show a steeper initial improvement, followed by a high-performance plateau for ratios above $\approx 0.5$. This indicates that, with more data, training yields a sufficiently powerful adapter representation, making performance robust to a wide range of higher $\alpha$ values. 

\section{Augmentation Analysis}

This analysis tests the possible addition of augmentations to HOSO-Adapter to make it more comparable with PathCLIP, which makes use of augmentations. In this approach, we create an additional, augmented view for each training image. The procedure applies weak and strong augmentations probabilistically, using policies such as RandAugment and RandomErasing. The original and the newly generated augmented views are passed to the model in a single forward pass, doubling the number of training views. 

Comparing Hoso-Adapter with augmentations (Table \ref{tab:rn50_augmentation_results}) to the main RN50 baseline yields minor differences: -0.54, -0.64, -0.20, and -0.43 percentage points for the 2-, 4-, 8-, and 16-shot settings, respectively. The introduced augmentations cause small performance drops; hence, we did not investigate augmentations further. 

\end{document}